\documentclass[letterpaper,11pt]{article}

\usepackage{a4wide}

\usepackage{mathptmx}

\usepackage{graphics}

\usepackage{natbib} 

\usepackage{tabularx}
\usepackage{amsmath}
\usepackage{amssymb}

\DeclareMathOperator*{\argmax}{arg\,max}

\DeclareMathOperator{\prob}{p}
\DeclareMathOperator{\expt}{E}
\def\tr{\text{tr}}

\begin{document}

\title{Random Forests on Distance Matrices for Imaging Genetics Studies}
\author{Aaron Sim, Dimosthenis Tsagkrasoulis and Giovanni Montana\footnote{to whom correspondence should be addressed: {\tt g.montana@imperial.ac.uk}}\\Statistics Section, Department of Mathematics\\Imperial College London}
\date{}

\maketitle

\begin{abstract}

We propose a non-parametric regression methodology, Random Forests on
Distance Matrices (RFDM), for detecting genetic variants associated to
quantitative phenotypes representing the human brain's structure or
function, and obtained using neuroimaging techniques. RFDM, which is an
extension of decision forests, requires a distance matrix as response that
encodes all pair-wise phenotypic distances in the random sample. We
discuss ways to learn such distances directly from the data using manifold
learning techniques, and how to define such distances when the
phenotypes are non-vectorial objects such as brain connectivity networks.
We also describe an extension of RFDM to detect espistatic effects while
keeping the computational complexity low. Extensive simulation results
and an application to an imaging genetics study of Alzheimer's Disease are
presented and discussed.

\end{abstract}

\section{Introduction}
{

The clinical pathology of neurological diseases and the imaging of the human brain are two areas of research that have largely developed along independent lines. It is only in the past few years that the usefulness of non-invasive imaging measurements of the human brain to the diagnosis and early prediction of neurological diseases been widely recognised \citep{albert, sperling, gray2}. In Alzheimer's Disease (AD), for instance, clinical guidance on the diagnosis of this most common of neurological degenerative disorders has recently been updated to incorporate neuroimaging markers alongside standard cognitive and behavioural tests \citep{albert, sperling}. The key to the improved characterisation of AD lies in the quantitative nature of the imaging measurements compared to the relatively subjective and imprecise nature of traditional clinical assessments. Imaging biomarkers of cerebral atrophy and of loss of connectivity between key regions in the brain are believed to be reliable indicators of AD and are particularly useful at early disease stages when standard cognitive assessments can be inconclusive. 

The utility of imaging phenotypes extends beyond diagnosis and prediction to the search for the underlying genetic factors behind neurological disorders \citep{mulm}. This comparatively more recent use of neuroimaging measurements in place of case-control labels in genetic association studies defines the emerging field of \textit{imaging genetics}. The central premise here is that, should they exist, genetic associations to intermediate brain structure and brain function phenotypes are stronger than those with the categorical clinical disease statuses further down the etiological chain \citep{Glahn2007}. Again, the example of AD serves as a good illustration. Although the key {APOE}$\varepsilon 4$ genetic variant was identified as a significant risk factor for AD almost 20 years ago \citep{apoe1, apoe2}, there has since been a general failure by numerous Genome-Wide Association Studies (GWAS) to uncover other genetic factors beyond a small set of variants. Because most GWAS rely on standard case-control status indicators (i.e. AD vs. cognitive normal) the lack of significant genetic findings suggests a deficit in statistical power inherent in such case-control experimental designs. More recently, by contrast, a number of new potential genetic variants have been reported \citep{silver, vounou2} in imaging genetics studies examining genetic factors which covary with phenotypes extracted from Magnetic Resonance Imaging (MRI) scans of the brain.

The effectiveness of imaging genetics is due largely to the inherently quantitative nature of imaging endophenotypes. In addition, neuroimaging data is often multivariate and, almost always, high-dimensional. For example, basic MRI scans provide high-dimensional brain-wide images of ventricular volumes at $\sim$1mm voxel resolutions; specialised MRI scans such as functional MRI (fMRI) and diffusion tensor imaging (DTI) provide a different set of similarly high-resolution phenotypes. fMRI measures the neural activity of brain regions via MR signals of the associated oxygenated blood flow while DTI maps the diffusion of water molecules, and hence the white matter fibre structure in brain tissue. One such phenotype arising from either fMRI time series data or from DTI maps of connected tissue structure is the brain connectivity network. Another common imaging modality is positron emission tomography (PET). With a range of radioactive tracers, PET provides non-volumetric, but similarly multivariate, quantitative phenotypes for imaging genetics. For instance, the fluorodeoxyglucose (FDG) tracer measures brain function through cerebral glucose metabolism at the voxel level, while the quantitative distribution of $\beta$-amyloid deposits can be ascertained using suitable tracers \citep{Drzezga2008}, the latter being implicated in AD pathology. Spatial correlations are extracted to generate a network view of the brain, while an analysis of a set of network images over time can provide a dynamical profile of loss of connectivity in patients with specific neurological diseases.

The simplest statistical approach to exploit the wealth of information in imaging genetics is known as Mass-Univariate Linear Modelling (MULM) \citep{vounou}. In MULM, multivariate phenotypes are treated as multiple univariate phenotypes, thereby sidestepping the complications inherent in multivariate analyses; for instance, where the number of single nucleotide polymorphism (SNP) covariates are typically several orders of magnitude higher than the number of subjects. This is, however, done at the expense of a reduction in power and an inability to detect interacting genetic effects. More recently, a number of multivariate approaches have been proposed for imaging genetics studies such as multi-locus penalised regression \citep{Kohannim2011} and sparse reduced-rank regression \citep{vounou}. Despite these developments, there are several features of imaging data not fully addressed and exploited by current methodologies. In this work we focus on three main aspects, as detailed below. 



First, although brain-wide imaging phenotypes can be very high-dimensional objects, the intrinsic dimensionality of the data is generally much smaller, and some form of dimensionality reduction prior to the imaging genetics study is commonly applied. In particular, it has been widely recognised that brain measurements lie on an embedded non-linear manifold within the higher-dimensional space \citep{gerber, Gerber:2009kx, Verma:2007vn}. In this reduced space, it may be easier to discriminate between populations of brains, for instance when healthy controls are compared to patients. Moreover, differences between these two populations are generally localised in specific brain regions, which may be detected by a suitable dimensionality reduction technique. For example, in AD patients, positive cerebral atrophy measurements are concentrated in the temporal lobe and, particularly, the hippocampus, amygdala and entorhinal cortex regions \citep{braak}. Manifold learning algorithms such as ISOMAP \citep{Tenenbaum:2000ys} and Laplacian eigenmaps \citep{laplace} are commonly used to infer the low-dimensional manifold from the observed data. In this paper we make use of a manifold estimated from a random sample of brains to compute a distance measure between any pair of brains, and then use such pair-wise distances, rather than the original measurements, to detect genetic associations.    

Second, some neuroimaging phenotypes may not fit well into a standard vectorial representation. For instance, an individual brain may be described by a covariance or correlation matrix specifying how the measurements at any two brain locations (e.g. regions of interest) co-vary in the brain. Individual brains can also be represented by their corresponding connectivity networks, where each network or graph specifies how any two brain regions are related to each other depending on whether they are connected in the network. The specific meaning of the presence or absence of a connection depends on the imaging modality (e.g. fMRI or DTI) and technique used to infer the network. In such cases, standard multivariate techniques to detect genetic associations with such quantitative phenotypes do not apply directly as they require vectorised measurements. Although it is in principle possible to vectorize a matrix as well as embed a graph into a vector, such simplifications would incur a loss of information, and are not ideal. Our proposed approach to deal with these cases is, again, to introduce a suitable distance measure between any two brain objects (e.g. matrices or networks), and then use such pair-wise distances to detect genetic associations.    

The third aspect we consider is the inability of integrating multiple quantitative endophenotypes. Several studies have demonstrated that integrating multiple phenotypes improves the overall performance in the classification of neurological diseases \citep{zhang, hinrichs}. Since we expect that this improvement carries over to genetic association studies, one would like to incorporate data from multiple modalities in imaging genetics studies. However, there is no natural method of combining multiple phenotypes using linear multivariate regression models. The phenotype representations might be different either due to different image modalities or simply different data structures (e.g. vectors, matrices and graphs). We argue that using pair-wise distances to represent differences in phenotypes which can then be explained by genetic variability provides a solution to this problem. 

In order to address the above aspects, in this paper we propose a non-parametric regression methodology, Random Forests on Distance Matrices (RFDM). RFDM is an extension of multivariate regression random forests whereby the response vector is replaced by a distance matrix. The elements of the distance matrix represent the pair-wise distances between quantitative phenotypes measures on subjects of a given experimental sample. The proposed distance-based approach is necessarily representation-independent and is, therefore, by construction, suited for all imaging modalities and data types. RFDM is complimentary to manifold learning techniques where the learned lower-dimensional manifold provides a natural metric to quantify all pairwise phenotypic distances between subjects. Furthermore, the use of distance matrices as responses in random forests provides a simple solution to the problem of combining imaging and other phenotypes by averaging distance or random forest proximity matrices.


Furthermore, the RFDM algorithm provides the means to progress beyond the detection of single genetic variants and their marginal effects to the detection of multi-way interactions between genetic markers, or epistatic effects. In the context of case-control GWAs, several methods have been developed for the detection of epistasis which do not resort to exhaustive searches yet retain acceptable levels of statistical power. Notable examples for categorical traits include multifactor dimensionality reduction \citep{Hahn2003}, the full interaction model \citep{Marchini2005}, Bayesian epistasis association mapping \citep{Zhang2007}, SNP harvester \citep{Yang2009}, and Maximum entropy conditional probability modelling \citep{miller} -- see \citet{chen} for a comparative analysis of these methods. More recently, a hierarchical maximum entropy modelling method for regression \citep{zhangHME} has been proposed to detect epistatic effects in the presence of univariate quantitative traits.

Random forests are particularly suitable for the detection of interacting features, and this aspect has been previously exploited in the context of GWAs \citep{McKinney2009, DeLobel2010, Jiang2009}. Typical strategies for detecting interacting SNPs include feature importance measures based on permutation schemes \citep{Bureau2005} or the selection of multiple features as node splitting criteria \citep{Yoshida2011}. In this paper we propose a computationally efficient information-theoretic epistasis detection methodology that works well in the RFDM framework, but could also be used in standard decision trees and forests. To the authors' knowledge, this is the first method to fully recognise and exploit multivariate quantitative phenotypes for the detection of epistatic interactions. 


In order to demonstrate how the proposed methodology can be used in imaging genetics studies we have performed an extensive number of simulation studies based on different genetic models and phenotypes. To this end we have devised an imaging genetics simulation model which relies upon a large simulated human population, and a disease model from which one can perform repeated sampling for multiple experiments. We show, through a simulated AD model, that RFDM significantly boosts the statistical power to detect causal SNPs, particularly when multiple datasets are integrated for the association testing. For non-vectorial imaging data, RFDM is shown to be highly effective, provided the distance matrix is suitable defined.

In addition to extensive simulation studies, we demonstrate the method through an application to a publicly dataset from the Alzheimer's Disease Neuroimaging Initiative (ADNI). Using a distance matrix to represent phenotypic differences at the neuroimaging level, DBRF is able to detect several well-known genetic variants associated to AD, namely the APOE$\varepsilon 4$ and TOMM40 genes. 


The rest of this paper is organised as follows. We begin in Section \ref{distRandomF} with an introduction to RFDM and provide a description of our distance-based generalisation. We also include a brief overview of the RF epistasis detection methodology. In Section \ref{distsection} we describe the processes of obtaining distance matrices. In Section \ref{admodeling} we outline the construction of an imaging genetics simulation engine for the purpose of validating of our proposed algorithm. We gather the details of the simulation set-up and the results of the experiments here. In Section \ref{realstudy} we provide the results of the study on ADNI data, and conclude with a discussion in Section \ref{summ}.

\section{Random Forests on Distance Matrices (RFDM)}\label{distRandomF}

\subsection{Decision forests with a generalised information gain} \label{dbgrf}
Consider a dataset $\mathsf{S}$ of $N$ feature-response pair observations
$
\mathsf{S} = \bigl\{(\mathbf{x}_i,\mathbf{y}_i)\,|\,\mathbf{x}_i\in \mathsf{X}, \,\mathbf{y}_i\in \mathsf{Y},\, i=1, \dotsc , N\bigr\}
$
where $\mathsf{X}$ is a $p$-dimensional feature space. We indicate with $\mathbf{x}_i = (x_{i1}, x_{i2}, \dotsc, x_{ip})$ the set of available SNPs for an individual where each element $x_{ij}$ represents the minor allele count at  locus $j$. We consider a range of phenotypic responses. For example $\mathbf{y}_i \in \mathbb{R}^n$ may represent the vector of voxel-wise brain volumes. The task at hand is to identify the SNPs that best predict the phenotypic trait $\mathbf{y}_i$. We adopt Random Forests for this regression problem. 

Random Forests (RF) is an ensemble machine learning tool which can be used for both feature selection and prediction \citep{breiman, CART, criminisi, eosl}. A forest consists of weakly correlated set of decision trees, each of which represents a distinct partition of the feature space. These partitions are constructed via greedy hierarchical processes where individual binary branchings are chosen to maximise a given objective function for corresponding subsets of the data. In this section we propose a novel objective function based on a generalised information gain using distance matrices; we refer to the overall methodology as Random Forest on Distance Matrices (RFDM).

At a given node $n$ of a given tree in the RF, we partition $\mathsf{X}$ via a binary and recursive splitting process by identifying a feature, indexed by $\alpha$, and split point $s$ to define a $(p-1)$-dimensional boundary $\{X\,|\,X\subset\mathsf{X}; \,X_\alpha = s\}$. This partitioning of $\mathsf{X}$ specifies a corresponding segmentation of $\mathsf{S}$ through the application of the splitting criteria ($x_{i\alpha}\leq s$ vs. $x_{i\alpha} > s$).  Let $S_n$, $S_{n\alpha s}^{(l)}$, and $S_{n\alpha s}^{(r)}$ represent the subsets of observations prior to splitting, and the post-split `left' $\{(\mathbf{x}_i, \mathbf{y}_i)\,|\, {x}_{i\alpha} \leq s\}$ and `right' $\{(\mathbf{x}_i, \mathbf{y}_i)\,|\, {x}_{i\alpha} > s\}$ branch subsets, respectively; let $I_n$, $I_{n\alpha s}^{(l)}$, and $I_{n\alpha s}^{(r)}$ be the corresponding sets of indices labelling the observations and $N_n$, $N_{n\alpha s}^{(l)}$, and $N_{n\alpha s}^{(r)}$ the number of observations in the respective subsets.

We place no restrictions on the response-space $\mathsf{Y}$ apart from the requirement that it is a metric space. Let $D$ be a symmetric, $N\times N$ positive real-valued matrix, where the components $D_{ij}$ denote the distances between $\mathbf{y}_i$ and $\mathbf{y}_j$ and satisfies the three required distance conditions $D_{ii} = 0$, $D_{ij} = D_{ji}$, and $D_{ij} + D_{jk} \geq D_{ik}, \forall\, i,j,k = 1, \dotsc, N$. We define the splitting objective function $G_{\alpha s}: \mathsf{P}(\mathsf{S}) \rightarrow \mathbb{R}$ as
\begin{equation}
G_{\alpha s}(S_n) = -\frac{1}{2N_n}\Biggl(\sum_{i,j\in l_n} \bigl(D_{ij}\bigr)^2 - \sum_{i,j\in I_{n\alpha s}^{(l)}}\bigl(D_{ij}\bigr)^2 - \sum_{i,j\in I_{n\alpha s}^{(r)}}\bigl(D_{ij}\bigr)^2\Biggr),\label{geninfogain}
\end{equation}
where $\mathsf{P}(\mathsf{S})$ is the power set of $\mathsf{S}$. The optimal splitting criteria $(\alpha_{nt}, s_{nt})$ at each node $n$ in tree $t$ is selected precisely to maximise the objective function resulting from the partition, i.e.
\begin{equation}
(\alpha_{nt},s_{nt}) = \argmax_{(\alpha,s)}G_{\alpha s}(S_n).\label{optcrit}
\end{equation}
As defined, $G_{\alpha s}$ generalises the RF regression and classification algorithm and dispenses with the dependence on the explicit representation of the responses $\mathbf{y}_i$. 

In a standard regression context, if $\mathbf{y}_i \in \mathbb{R}^q$, then one simply recovers the Euclidean distances $D_{ij} = |\mathbf{y}_i - \mathbf{y}_j| \equiv [(\mathbf{y}_i - \mathbf{y}_j)^T(\mathbf{y}_i - \mathbf{y}_j)]^{1/2}$. Using the well-known equivalence relation expressing the second central moment in terms of sum over all squared pair-wise distances, i.e.
\begin{equation}
\sum_{i=1}^N|\mathbf{y}_i-\bar{\mathbf{y}}|^2 = \frac{1}{2N}\sum_{i=1}^N\sum_{j=1}^N|\mathbf{y}_i - \mathbf{y}_j|^2,\label{equi}
\end{equation}
we obtain the familiar regression objective function
\begin{equation}
G_{\alpha s}^E(S_n) = \sum_{i\in I_n}|\mathbf{y}_i - \bar{\mathbf{y}}_n|^2 - \sum_{i\in I_n^{(l)}}|\mathbf{y}_i - \bar{\mathbf{y}}_{n\alpha s}^{(l)}|^2 - \sum_{i\in I_n^{(r)}}|\mathbf{y}_i - \bar{\mathbf{y}}_{n\alpha s}^{(r)}|^2,\label{regsplitcrit}
\end{equation}
where $\bar{\mathbf{y}}_n$, $\bar{\mathbf{y}}_{n\alpha s}^{(l)}$, and $\bar{\mathbf{y}}_{n\alpha s}^{(r)}$ are the Euclidean mean response vectors of the data subsets in node $n$ and its left and right child nodes respectively. If we further assume that the deviation of the data about the means are normally distributed with some fixed diagonal variance-covariance $\Sigma = \mathbf{1}_q \sigma^2$, i.e.
\begin{equation}
\log p(\mathbf{y}) \propto (\mathbf{y} - \bar{\mathbf{y}})^2,
\end{equation}
we have, for the individual terms in \eqref{regsplitcrit}, an equivalence to the Shannon entropy\footnote{The overall scaling factor and constant additive term arising from the omitted variance terms in \eqref{ssrinfo} can be ignored as only the \textit{relative} values of $G_{n\alpha s}$ are of interest.}
\begin{equation}\label{ssrinfo}
\sum_\mathbf{y}(\mathbf{y} - \bar{\mathbf{y}})^2  = \expt_p \bigl[(\mathbf{y} - \bar{\mathbf{y}})^2\bigr] = -\kappa\expt_p\bigl[\log p(\mathbf{y})\bigr] = -\kappa\sum_{\mathbf{y}}p(\mathbf{y})\log p(\mathbf{y})
\end{equation}
where $\kappa = \kappa(\sigma)$ is a constant for fixed $\sigma$. In this case, the regression objective function \eqref{regsplitcrit} is a measure of information gain; in the representation-free case, $G_{\alpha s}$ \eqref{geninfogain} can, therefore, be loosely interpreted as a \textit{generalised} information gain measure, defined for the cases when the response mean $\bar{\mathbf{y}}$ is not defined. 

Classification forests can also be seen as special cases. In this context, $\mathsf{Y}$ has a discrete topology with $c$ points corresponding to $c$ categories. We define a discrete metric on the space where distances are given by 
\begin{equation}
D_{ij} = 
\begin{cases}
    0,& \,\,\text{if } \mathbf{y}_i = \mathbf{y}_j\\
    1,              &\,\, \text{otherwise}
\end{cases}.
\end{equation}
The distance measure $I = -\frac{1}{2N}\sum_{i,j=1}^N(D_{ij})^2$ in \eqref{geninfogain} has the expected behaviour of an information measure. For instance, we have the maximum
$I = 0$ if $\mathbf{y}_i = \mathbf{y}_j$ for all $i, j$; similarly $I$ is minimised for when the off-diagonal terms in $D$ are maximised, i.e. the conventional entropy is greatest with data evenly distributed across categories. Nevertheless, for classification, the generalised information gain via distance matrices $D$ is not strictly necessary as the conventional objective function is never ill-defined. The Shannon information gain for categorical responses is simply
\begin{equation}
G^{\text{Sh}}_{\alpha s}(S_n) = -\sum_{\mathbf{y}\in S_n}p_n(\mathbf{y}) \log p_n(\mathbf{y}) + \sum_{\mathbf{y}\in S_{n\alpha s}^{(l)}}p_n^{(l)}(\mathbf{y}) \log p_n^{(l)}(\mathbf{y}) + \sum_{\mathbf{y}\in S_{n\alpha s}^{(r)}}p_n^{(r)}(\mathbf{y}) \log p_n^{(r)}(\mathbf{y}),
\end{equation}
where
\begin{equation}
p_n(\mathbf{y}) = \frac{1}{N_n}\sum_{i\in I_n}\mathbb{I}_{\mathbf{y}}(\mathbf{y}_i), \quad p_n^{(l)}(\mathbf{y}) =\frac{1}{N_{n\alpha s}^{(l)}}\sum_{i\in I_{n\alpha s}^{(l)}}\mathbb{I}_{\mathbf{y}}(\mathbf{y}_i), \quad p_n^{(r)}(\mathbf{y}) =\frac{1}{N_{n\alpha s}^{(r)}}\sum_{i\in I_{n\alpha s}^{(r)}}\mathbb{I}_{\mathbf{y}}(\mathbf{y}_i),
\end{equation} 
are the respective probability mass functions, with the indicator function $\mathbb{I}_{\mathbf{y_i}}(\mathbf{y}_j) = \delta_{ij}$. 

In many cases\footnote{Exceptions include responses with infinite degrees of freedom, where the forced vectorial representations are infinite-dimensional; e.g. functions represented by its infinite vector of Fourier modes.}, as is quite common in the practice, one can avoid the need for RFDM by forcing the responses into vectors. For example, a graph may be represented by the vector of components in its adjacency matrix. However, given the arbitrary nature of such constructions, a naive application of a flat Euclidean norm in calculating $\bar{\mathbf{y}}$ is likely to be just as arbitrary.

The removal of the dependence on the explicit representation of the responses provides a potential avenue for integrating data from various sources through the construction of a weighted average over distance measures $D_{ij}^{(1)}, D_{ij}^{(2)}, \dotsc$, i.e.
\begin{equation}\label{distcomb}
\overline{D}_{ij} = \sum_{a=1}^{a_{\text{max}}}w_a D_{ij}^{(a)}, \quad\text{subject to}\quad \sum_{a=1}^{a_{\text{max}}} w_a = 1.
\end{equation} 
In the context of imaging genetics, each $D^{(a)}$ represents a different imaging signature of the brain. In practice, however, the attempt to integrate data in this manner subtly reintroduces the representation dependency that one might be seeking to dispense with. The various distances $D^{(a)}$ might either be scaled differently with (often unknown) non-linear relationships, or have different domains altogether, e.g. $D_{ij}\in [0, \infty)$ vs. $[0,1)$. In both instances it is not obvious how to average distances. For trivial cases where the representations are identical\footnote{For example, identical measurements taken at different time points.}, \eqref{distcomb} can be used to combine the data. In the general case, one requires a mapping from $D_{ij}^{(a)}$ to a common representation in which different modalities can be integrated.

\subsection{Gini variable importance measures for SNP ranking and epistasis detection}

In addition to phenotype prediction, the main utility of RF in genetic association studies lies in its ability to rank SNPs according to their marginal importances. See \citep{goldstein} for a review. 

 
In this paper we use the Gini variable importance measure for feature selection. The Gini information score $G_\alpha$ for a SNP indexed by $\alpha$ is
\begin{equation}\label{giniindex}
G_\alpha = \frac{1}{g_\alpha}\sum_{t=1}^T\sum_{n\in M_t} G_{tn} \mathbb{I}_\alpha (\alpha_{tn}),
\end{equation}
where $M_t$ is the set of node indices in tree $t$ and the normalisation factor $g_\alpha$ is the total number of nodes across every tree in the RF where the SNP $\alpha$ was a candidate in the splitting criteria search.

\begin{figure}[ht] 
\centering
{\includegraphics{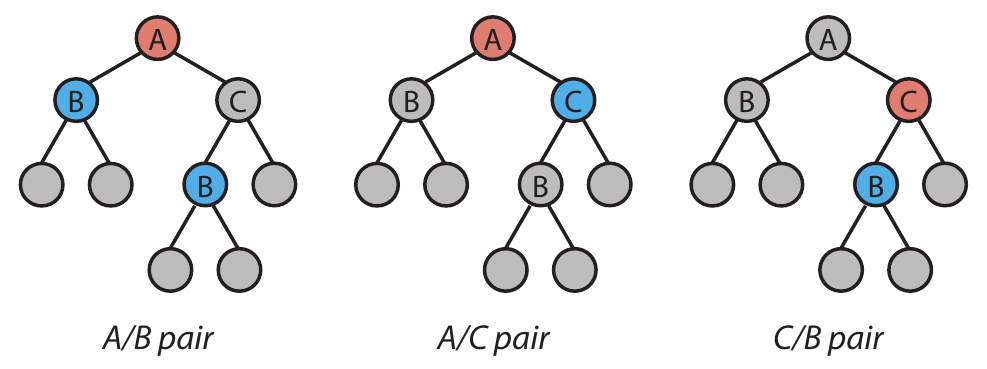}}
\caption[Gini pairwise interaction example]{Illustration of the variable pairings for the Gini pairwise interaction measure in a single, highly-simplified, tree. The labels on each node indicate the variables (SNPs) selected to perform the data partitioning; nodes without labels are terminal nodes. The blue colours indicate the nodes where the information gain is evaluated for the calculation of the Gini pairwise interaction measure. The Gini pairwise measure, in each instance corresponding to a specific feature pair, is the absolute difference between the Gini indices of the left and right branches (blue vertices) of the parent node (red vertices), normalised by the number of data points in the parent node. The instances are summed over all possible parent nodes that have at least one non-terminal child node, across every tree in the RF.} 
\label{giniinteract}
\end{figure}

The ability of RF to capture epistatic effects is well-documented \citep{sun} and can be understood intuitively. If a variable is chosen for the splitting criteria \eqref{optcrit} on one branch of a tree but not another, there is an increased likelihood that the variable interacts with the variable which produced the original branching. Specifically, an interaction between two features manifests as an inequality between the \textit{conditional} and \textit{marginal} variable importances. The disadvantage of using the standard RF methodology to detect interactions is that whilst the various variable ranking methods rank the full marginal effects (which includes, if present, interacting effects), it does not identify the interacting pairs or higher-order combinations. So a highly interacting feature may have a high variable importance score but its interacting partner remains unspecified. A simplistic solution is to consider the space of all possible feature combinations and to identify the interacting sets via the standard Gini variable importance measure. However this method, even when restricted to pair-wise interactions in moderately high-dimensional set-ups, is computationally prohibitive.

We propose a measure to precisely quantify, and therefore rank, all pairwise interaction measures from a RF run. It is based on a normalised sum of the left/right branch conditional Gini information measure differences \eqref{giniindex} over every sub-trees in each decision tree. We refer to this new criterion as the \textit{Gini pairwise interaction measure}. For two SNPs indexed by $\alpha$ and $\beta$, we define the conditional Gini importance $G_{\alpha\beta}$ as
\begin{equation}
G_{\alpha\beta} \equiv G_{\alpha|\beta} + G_{\beta|\alpha},
\end{equation}
where
\begin{equation}\label{ginipairindex}
G_{\alpha|\beta} := \frac{1}{2}\sum_{t=1}^{T}\sum_{n\in M_t} \mathbb{I}_{\alpha} (\alpha_{tn}) \Biggl[\sum_{n\in M_{tn}^{(l)}} \frac{N_n}{N_n^{(l)}}\mathbb{I}_{\beta}(\beta_{tn}) G_{tn} - \sum_{n\in M_{tn}^{(r)}} \frac{N_n}{N_n^{(r)}}\mathbb{I}_{\beta}(\beta_{tn}) G_{tn}\Biggr]^2,
\end{equation}
with $M_{tn}^{(l,r)}$ the sets of node indices contained in the left and right branches, respectively, from node $n$ in tree $t$. For a given node split on the feature $\alpha$, the terms in the parentheses in \eqref{ginipairindex} give the difference between the left and right sum of Gini distance information scores of all the subsequent nodes split on feature $\beta$. The sums across the branches are normalised according to the number of data points in the respective child nodes prior to taking the difference. This quantity is calculated for and summed across all nodes in the tree split on $\alpha$, and across all trees in the RF. 

We illustrate the proposed measure for a simple toy example in Figure \ref{giniinteract}. We have three features, labelled A, B, and C, and consider only one, three-level deep, tree in the RF. We also assume that all the features are considered in every node splitting decision. In each tree figure we highlight the calculation of the conditional Gini importance of a certain feature pair. The interactions are ranked according to the magnitude (not illustrated) of the difference in importance scores. In this simple example, each feature pair has at most one associated diagram; for larger, more realistic, trees, there may be several tree diagrams for each feature pair.

Apart from identifying specific pairings, the key advantage of this novel approach to epistasis detection is that no significant additional computational effort is required beyond the original RF run. This is because the Gini pairwise interaction measures are completely determined from the Gini importance measures and the tree structures in the RF; the only computational overhead incurred involves the identification of the relevant feature pairs across the RF -- a relatively cheap cost. This is in contrast with the highly expensive brute force strategy of running a RF for the space of all feature pairs, or calculating the pairwise permutation variable importances.


\section{Distance metrics for neuroimaging phenotypes}\label{distsection}
The two motivating advantages of RFDM in imaging genetics are to incorporate non-vectorial imaging responses, and to integrate data across multiple modalities. In this section we turn to the task of selecting the metrics or methods through which one obtains the distance matrices $D_{ij}^{(a)}$, focusing on the specific neuroimaging phenotypes encountered.

\subsection{Manifold learning approaches}\label{manifoldforests}
Manifold learning techniques seek to uncover from the original data a lower dimensional submanifold of the full high-dimensional data space. These submanifolds contain, in an approximate manner, the original data points and preserves \textit{local} pairwise Euclidean distances. They are typically presented in terms of coordinate charts that are constructed such that Euclidean pairwise distances equal the respective geodesic path lengths on the submanifold. 

The manifold learning approach in RFDM constructs the distance matrix using these submanifold pairwise distances. This is in contrast to simply employing the Euclidean metric in the full high-dimensional space. The validity of the latter procedure implicitly assumes the existence of unobserved (out-of-sample) data points in the interpolating regions between every pair of points in the original data set. It has been shown that this assumption is not valid for brain images and hence the current trend of using machine learning to uncover the manifold structure of neuroimages emerged \citep{gerber}.

In this paper we will adopt two such techniques. The first, recently employed in several neuroimaging applications \citep{gray, gerber}, is Laplacian eigenmaps \citep{laplace} from spectral graph theory. This allows one to translate a similarity matrix of a graphical representation into a distance matrix. The second is Totally Random Trees Embedding (TRTE) \citep{Pedregosa2011}, which relies on learning decision trees.

\subsection{Predefined distances}

For well-studied data representations, there are often several analytical expressions for the distance metric, whose properties are known, from which one can choose for the problem at hand \citep{minas, deza}. The two non-vectorial neuroimaging phenotype representations we consider are brain connectivity networks and brain covariance matrices.

The brain connectivity network of a single subject can either be inferred from spatial correlations in brain chemical activity (as in PET and fMRI) or from structural observations of the brain (DTI). These networks are represented by undirected graphs with different brain regions as nodes, with edges that can be weighted; in the case of PET imaging, for example, the weights are scaled to the strength of the correlations between the corresponding brain regions in longitudinal scan data. 

Performing a regression on networks in RFDM require one to determine the distance \textit{between} networks. There are several metrics defined on the space of graphs \citep{deza}. Since the vertices are identical for each separate brain network, the metric is simply a function of the graph edges representing the connectivity between brain regions. Since the severity of certain neurological diseases (e.g. AD) is described by a simple monotonic decline in brain connectivity, as detailed in Appendix \ref{endosim}, we define the distance between two isomorphic, vertex-labelled graphs $G_i$ and $G_j$ by the absolute difference
\begin{equation}\label{graphdist}
D_{ij} = |E_i -  E_j|,
\end{equation}
where $E_i$ is the number of edges in graph $G_i$; the symmetry, positivity and triangle inequality constraints are trivially satisfied by \eqref{graphdist}.

Brain covariance matrices of a single subject can either be lifted directly from PET and fMRI spatial correlations, or inferred from correlations in longitudinal basic MRI scans. In Appendix \ref{endosim} we outline the simulation of covariance matrices from longitudinal volumetric MRI data. As shown in \citet{covariance}, a candidate distance between two symmetric, positive definite, matrices $A_i$ and $A_j$ is given by
\begin{equation}\label{covdist}
D_{ij} = \biggl[\sum_{a=1}^n\log^2 \lambda_a(A_i, A_j)\biggr]^{\frac{1}{2}},
\end{equation}
where the generalised eigenvalues $\lambda_a(A_i, A_j)$ are defined via $|\lambda A_i - A_j| = 0$.

\subsection{Phenotype data integration through distances}

Integrating datasets across different modalities is one way to boost the statistical power of genetic association tests relative to those performed separately with the individual datasets. Evidence for this lies in the demonstrated improvement in predictive power. One method used to integrate the data, ironically, makes use of the supposedly unreliable case-control labels.

Consider an augmented dataset $\mathsf{S}'$ of $N$ observations
\begin{equation}\label{augmented_s}
\mathsf{S}' = \bigl\{(\mathbf{x}_i,\mathbf{y}_i, \mathbf{z}_i)\,|\,\mathbf{x}_i\in \mathsf{X}, \,\mathbf{y}_i\in \mathsf{Y}, \,\mathbf{z}_i\in \mathsf{Z},\, i=1, \dotsc , N\bigr\}.
\end{equation}
where, in addition to the features $\mathbf{x}_i$ and the intermediate multivariate quantitative response $\mathbf{y}_i$, $\mathbf{z}_i$ is an additional response variable dependent on $\mathbf{y}_i$, i.e. $\mathbf{z}_i = f(\mathbf{y}_i)$; $\mathbf{z}_i$ is typically a binary response variable, i.e. $\mathsf{Z} = \mathbb{Z}^2$; the case/control classification is the canonical example. From a RF classification run with $(\mathbf{z}_i, \mathbf{y}_i)$ as the response-feature variable pair, with the components of $\mathbf{y}_i$ as the regression covariates, we obtain the $(N\times N)$-dimensional proximity matrix $W$ between the response variables as a normalised count of the number of co-occurrence of observations at the terminal nodes of the RF trees, i.e.

\begin{equation}\label{proximity}
W_{ij} = \frac{1}{2w_{ij}}\sum_{t=1}^{T}\sum_{n\in {M}'_t}\sum_{k,l\in I_n} \mathbb{I}_{(i,j)} (k,l),
\end{equation}
where normalising factor $w_{ij}$ is equal to the number of OOB sets in $\{S_{1}^{\bot}, S_{2}^{\bot}, \dotsc, S_{T}^{\bot}\}$ containing both observations $i$ and $j$, and $M'_t$ is the set of terminal node indices of tree $t$ in the RF.


With the manifold learning procedure, there are two inequivalent ways of integrating data from different modalities. We can either evaluate the mean of the distances \eqref{distcomb}, or perform the integration at the level of the proximity matrices themselves. Following \citep{gray}, we adopt the latter option. Given the proximity matrices $W_{ij}^{(1)}, W_{ij}^{(2)}, \dotsc$, we define the weighted average proximity matrix
\begin{equation}\label{proxcomb}
\overline{W}_{ij} = \sum_{a=1}^{a_{\text{max}}}w_a W_{ij}^{(a)}, \quad\text{subject to}\quad \sum_{a=1}^{a_{\text{max}}} w_a = 1.
\end{equation} 
The integrated distance matrix $\overline{D}_{ij}$ is then determined from $\overline{W}_{ij}$ using Laplacian eigenmaps, as described above.


\section{Simulation studies}\label{admodeling}
In this section we outline the framework of an imaging genetics simulation model that we have used to validate RFDM. Full details of the simulation engine, including the real datasets used to ensure an appropriate level of realism, are given in Appendix \ref{simdetails}. We then describe the simulations and present the results.

\subsection{An imaging genetics simulation engine}

Our simulation is similar to the approach taken in \citet{vounou} and is decomposed to three parts, which are described below. In the first stage, we construct a base population with a realistically simulated genotype distribution and several quantitative imaging phenotypes. There is no constraint on either the number or the types of phenotypes. In the second stage, we propose a two-step disease model which first dictates, according to the genotype, modifications to the base phenotypes, before providing a disease classification as a function of the modified quantitative phenotype. This classification step serves the following two purposes. Firstly it allows us to establish the power of imaging genetics via a comparison of quantitative vs. case-control RF regressions, and secondly it is used to validate the supervised learning approach to define distances between brain networks. Finally, we apply our method to a set of random samples from the simulated population and assess its ability to identify the genetic causes using receiver operating characteristic (ROC) curves. The model is represented diagrammatically in Figure \ref{model}.

\begin{figure}[ht] 
\centering
{\includegraphics{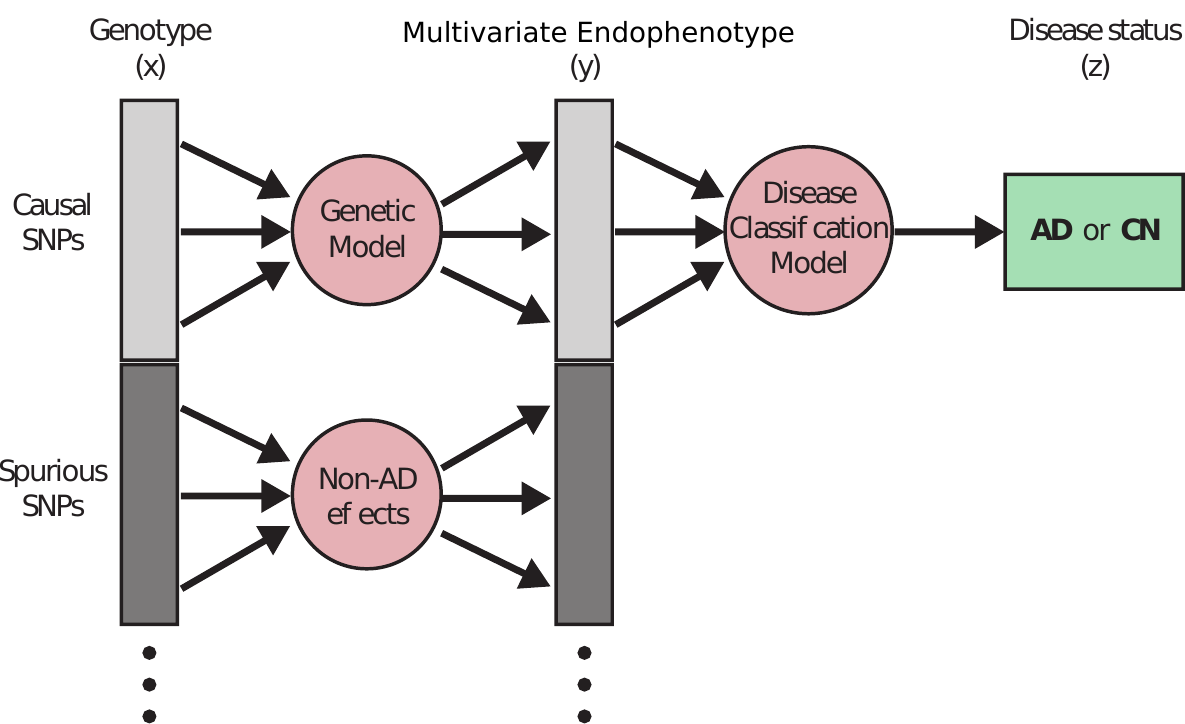}}
\caption[Disease model]{The simulated genetics effects and disease classification model. The arrows represent a causal effect. The inclusion of `spurious' SNPs depends on the particular simulation performed.} 
\label{model}
\end{figure}

We focus on the detection of epistasis in the possible presence of other single loci effects. To demonstrate the applicability of our RF algorithm across multiple scenarios, we consider four different epistasis models, which we label $P_1$ to $P_4$.  $P_1$, $P_2$, and $P_3$ contain only pair-wise interaction effects, while $P_4$ contains both interaction and non-interaction effects. In addition to the interaction models, for the purposes of demonstrating the standard application of RF to detect predictor variable importances, we consider a non-interacting model, $P_0$, where there are only non-interacting SNPs; in this case only do we test for the detection of individual SNPs rather than interacting SNP pairs. The five sets of causal SNPs and SNP pairs are
\begin{align}
P_0 &= \{ \alpha_1, \alpha_2, \dotsc , \alpha_7\}  & \text{(Non-interacting)},\nonumber\\
P_1 &= \{({\alpha_1}, {\alpha_2}), ({\alpha_3}, {\alpha_4}), \dotsc ({\alpha_{15}}, {\alpha_{16}})\}  & \text{(Non-overlapping pairs)},\nonumber\\
P_2 &= \{({\alpha_1}, {\alpha_2}), ({\alpha_1}, {\alpha_3}), \dotsc ({\alpha_{1}}, {\alpha_{8}})\}  & \text{(Central node)},\nonumber\\
P_3 &= \{({\alpha_1}, {\alpha_2}), ({\alpha_2}, {\alpha_3}), \dotsc ({\alpha_{8}}, {\alpha_{9}})\} & \text{(Chain-linked)} ,\nonumber \\
P_4 &= \{({\alpha_1}, {\alpha_2}), ({\alpha_2}, {\alpha_3}), \dotsc ({\alpha_{8}}, {\alpha_{9}}), {\alpha_{10}}, \dotsc, {\alpha_{16}}\}& \text{(Chain-linked, mixed model)}, \nonumber
\end{align}
where $\alpha_\cdot\in S_c$ are the feature indices. In $P_1$ to $P_4$ we assume eight causal SNP pairs, chosen at random from $S_{\text{c}}$. The epistasis models are summarised in Figure \ref{epimodels}.

\begin{figure}[ht] 
\centering
\begin{tabular}{c}
{\includegraphics{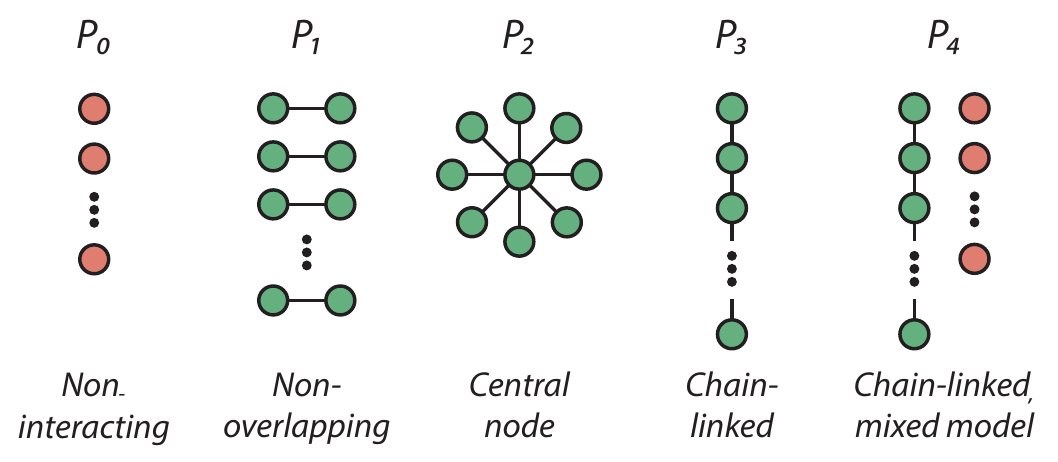}}
\end{tabular}
\caption[Genetics models]{The five genetic models. The green circles indicate SNPs with purely interaction effects; the red circles indicate SNPs with only non-interacting, purely marginal, effects. All four models contain eight interacting pairs, with possibly one or more SNPs interacting in multiple pairs. In each of models $P_0$ and $P_4$, there are seven non-interacting SNPs.}
\label{epimodels}
\end{figure}

\subsection{Simulation results}\label{simresults}

In our simulation study we consider three imaging phenotypes which we represent by $\mathbf{y}_i$, $\Sigma^{(i)}$ and $G_i$, with $i$ the subject index. The first phenotype is the rate of volume change across the brain and is represented by a vector $\mathbf{y}_i$ of time-averaged expansion rates where each component corresponds to a given brain region. The second phenotype is the brain connectivity network represented by the graph adjacency matrix $\Sigma^{(i)}$ . The third phenotype is the raw covariance matrix representing the covarying voxel volume measurements. These are given by the square matrix $G_i$. The respective genetic-related phenotypic modifications are an increased rate of brain atrophy and an attenuation in brain network connectivity, the latter manifesting either by a reduction in inter-region correlations or loss of graph connections. We adopt a binary disease classification model, i.e. case-control. Full details of the simulation of the three phenotypes are presented in Appendix \ref{endosim}.

In our simulation experiments, the RFs are built with 500 trees. We adopt standard parameters, with $\textit{mtry} = \sqrt{p}$ for classification and $\textit{mtry} = p/3$ for regression trees, where $p$ is the total number of features. Following the multivariate RF analysis in \citep{segal}, we impose a maximum tree depth $d=7$. Each population sample contains 200 subjects, with equal proportions of cases and controls.

We begin by verifying both the utility of quantitative phenotypes over simple case/control classifications. We then validate our proposed RF epistasis detection methodology for four genetic effects models described above (P1 -- P4). Following these preliminary tests, we demonstrate the power of the RFDM over standard, Euclidean-based, multivariate RF analyses in three generalised distance simulations -- incorporating the presence of non-vectorial phenotypes, combining modalities, and dealing with spurious signals in the endophenotypes. In each experiment, we plot ROC curves averaged over multiple iterations, wherein different random samples of causal SNPs and disease-linked brain ROIs are selected. 
 
\subsubsection{Comparing case-control with quantitative phenotypes}

\paragraph{Marginal SNP detection}

The goals of this experiment are to demonstrate the use of RF is SNP detection and to validate the claim of imaging genetics, that the use of quantitative phenotypes increases the statistical power to detect causal SNPs, relative to the use of case-control statuses. We set $P_0$ to be the causal-SNPs set (7 causal SNPs with no interactions). The effect-size parameter is fixed\footnote{This value is chosen to ensure a measurable but weak signal of causal SNPs in the case-control set-up} as $\delta=1.5$. We modify, via the genetic model \eqref{volmod}, the base volume gradient $y$ and generate our sample sets using the disease classification model \eqref{bayespos}. By adjusting the tuning parameter $\zeta$ in \eqref{tuning}, we set the penetrance level of the disease in the population at 20\%. Finally, using either the case-control labels or quantitative phenotype vector $y^*$ as our response variable, we perform a RF classification and a multivariate RF regression respectively. 

As shown in Figure \ref{imggen}, the use of quantitative endophenotypes as responses outperforms the set-up with case-control responses. For the purposes of identifying the SNPs with the strongest effects, it is clear that the use of quantitative phenotypes allows one to identify a significant proportion of causal SNPs unlike similar searches using case-control labels.

\begin{figure}[ht] 
\centering
{\includegraphics{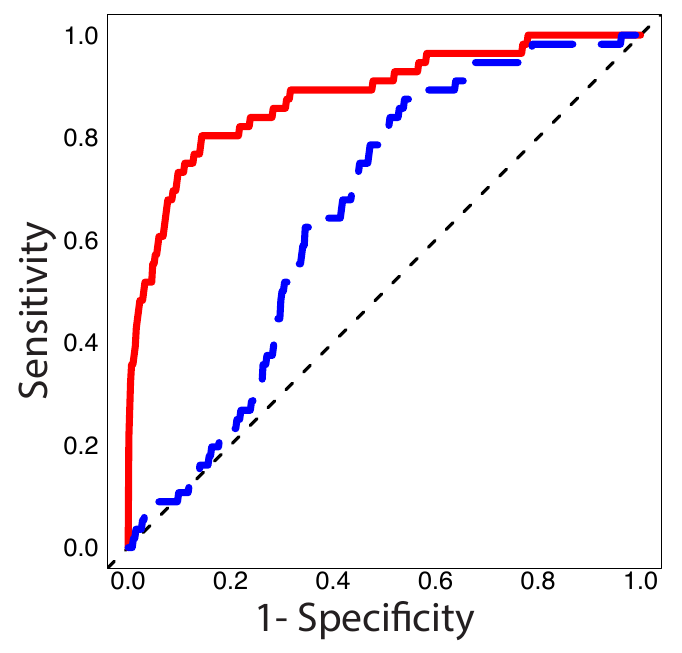}} 
\caption[Quantitative vs. categorical power]{Mean ROC curves for the detection of individual causal SNPs (as simulated without any SNP-SNP interactions), as averaged over 8 RF runs, for population disease penetrance = 20\%. The blue dashed curves represent the mean output from RF classifications with case-control responses; the red solid curves represent the mean output from RF regression with quantitative, multivariate, responses.}
\label{imggen}
\end{figure}

\paragraph{Epistasis detection}
The purpose of this experiment is to demonstrate the utility of RF in detecting epistatic effects for a variety of genetic models.  The causal-SNPs and SNP-pairs are given by four model sets $P_1$--$P_4$. Identically to Experiment 1 above, we adopt the effect size $\delta=1.5$. Here we consider two penetrance levels of 0.2 and 0.35, and perform the RF runs using both the quantitative volumetric gradients and case-control labels. Instead of the Gini information score for single variables, we rank the 73920 ($(385\times 384)/2$) SNP-SNP pairs using the Gini pairwise interaction measure \eqref{ginipairindex}. 

We make three observations from the resulting ROC curves in Figure \ref{imggen2}. Firstly, it is clear that the RF conditional Gini information is an effective measure for the identification of specific interacting SNP-SNP pairs for a variety of interaction models, including those with the inclusion of non-interacting SNP effects. Secondly, similar to the detection of marginal effects (Experiment 1), the use of quantitative phenotypes increases the statistical power of association studies. Thirdly, despite the efficacy in detecting interactions, the conditional Gini information is not `equitable'  in the sense that detection power varies with different genetic models. This latter property is easily explained. In situations where one or more SNPs in causal SNP-SNP pairs have strong marginal effects (the central node SNP in model $P_2$, for example), there is a higher likelihood that the selection of those SNPs as the node-splitting features occur `higher up the trees' where the sample sizes are larger and the detection power greater.

\begin{figure}[ht] 
\centering
{\includegraphics{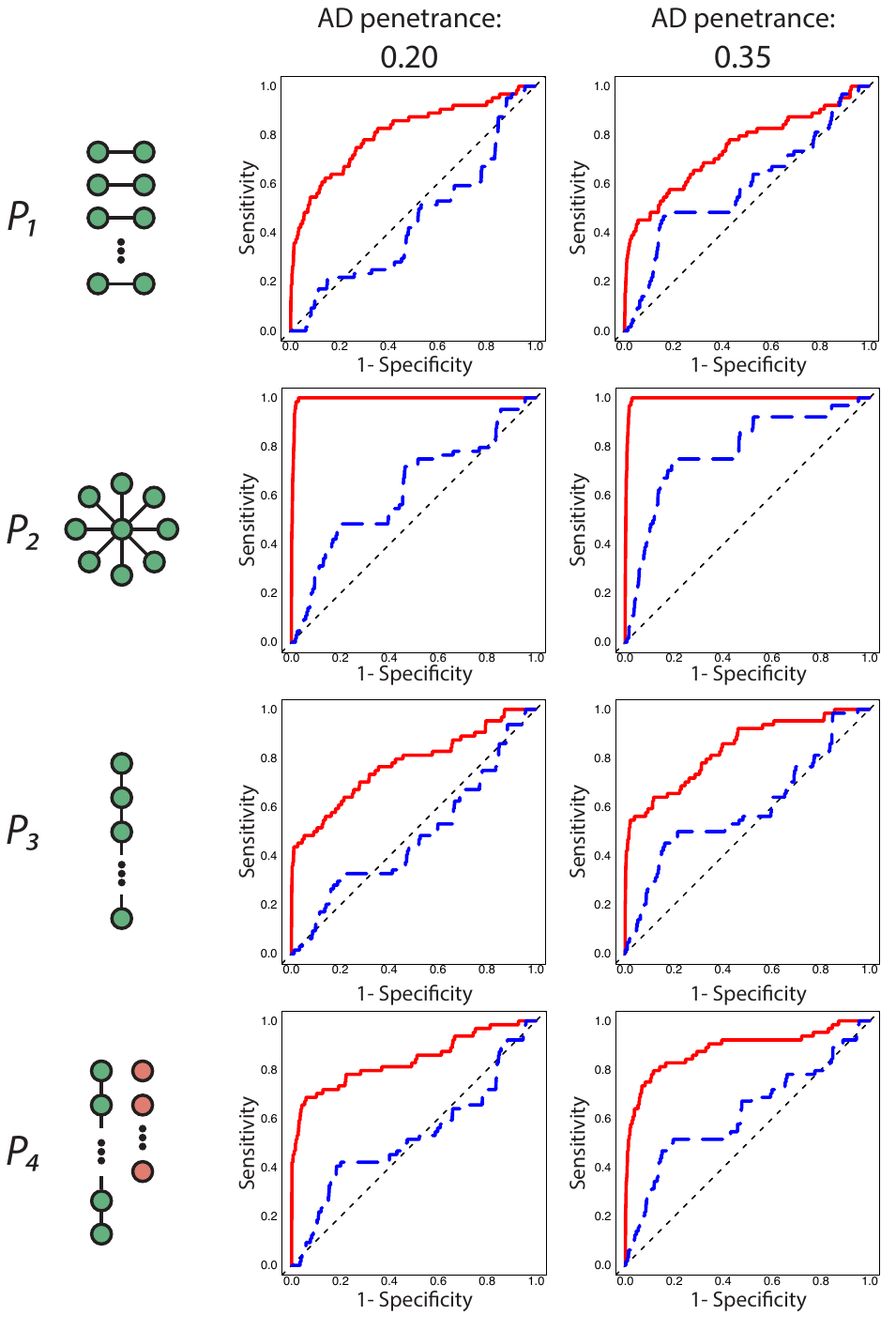}} 
\caption[Epistasis experiment]{Mean ROC curves for the detection of causal SNP-SNP pairs, as averaged over 8 RF runs. \textbf{Left column}: population disease penetrance = 0.20; \textbf{Right column}: population disease penetrance = 0.35. The blue dashed curves represent the mean output from RF classifications with case-control responses; the red solid curves represent the mean output from RF regression with multivariate responses. The four rows represent, in order, the four different epistasis disease models $P_1 - P_4$.}
\label{imggen2}
\end{figure}

\subsubsection{Generalised distance simulations}
In the previous two experiments, the quantitative phenotype was treated as a Euclidean vector. Specifically, the Euclidean norm was used in the evaluation of the splitting criteria \eqref{equi}. In the three subsequent experiments, we simulate scenarios where this default treatment of a quantitative response is, respectively, either impossible, limiting, or misleading. The main task in generalising to a distance-based approach is to derive the distance matrix $D_{ij}$; once defined, the adoption of the generalised splitting criteria \eqref{geninfogain} is relatively straightforward, as outlined in Section \ref{dbgrf}. We adopt a population disease penetrance level of 20\% throughout, restrict ourselves to the interaction model $P_3$, and, where applicable and unless otherwise stated, set the effect size on the brain volume vector to be $\delta=1.5$. 

\paragraph{Covariance matrix phenotype}
In this experiment we simulate non-vectorial quantitative phenotypes -- covariance matrices representing brain connectivity and demonstrate the use of specific, non-Euclidean, distance measures to derive the distance matrices for use in a RF regression runs. 

Starting from the assignment of base covariance matrices $\{\Sigma^{(i)}_{kl}\}$ \eqref{basecov} representing the brain connectivity between the 400 ROI, we implement the genetic model attenuation \eqref{covmodel} with effect size $\gamma=0.3$.  The distance matrices $D^{\Sigma}_{ij}$ is obtained from \eqref{covdist}. We perform the RFDM regression and compare the output to the RF classification using case-control labels in Experiment 2.

The results of this and the next experiment are presented together in Figure \ref{epimodel3}.

\paragraph{Connectivity network phenotype}
From the covariance matrices in the previous experiment, we infer a set of undirected graphs $\{G_i\}$ using sparse inverse covariance estimation (SICE) \eqref{SICEmax}. The distance matrices $D^{G}_{ij}$ are obtained using \eqref{graphdist}.

The results are shown in Figure \ref{epimodel3} (subplot A). We see that, of the two distance metrics, it is the graph metric $D^{G}_{ij}$ which gives the best results; $D^\Sigma_{ij}$, despite improving on the case-control association test, is a relatively poor choice due to the lack of robustness of the high-dimensional covariance matrices. One can determine the underlying reason for the relative shapes of the ROC curves by examining the manifold plots, shown in Figure \ref{epimodel3}, obtained from the proximity matrices of the two RFDM regression runs. We observe that the clustering implied by employing the graph metric is significantly more pronounced than the one implied by the use of $D^\Sigma_{ij}$. We can infer that the graph induced clustering (subplot C) is itself a partition for the minor allele frequencies of the causal SNPs. We give an explicit illustration of this last point in Experiment 4 below.

For illustrative purposes, we select four subjects from the manifold plot in Figure \ref{epimodel3} (subplot C) and plot their SICE-inferred graphs. This is shown in Figure \ref{graphplots}. As expected from the model of connectivity loss in AD patients, the relative density of the graphs $\{a,b\}$ in the top cluster compared to the graphs $\{c,d\}$ in the bottom cluster coincides with the greater proportion of disease subjects in the former. From this one illustrative example, we can deduce that the use of specific, well-chosen, non-Euclidean distance measures enables one to harness the statistical power from quantitative but non-vectorial phenotypes in genetic association studies. 


\begin{figure}[ht] 
\centering
\includegraphics{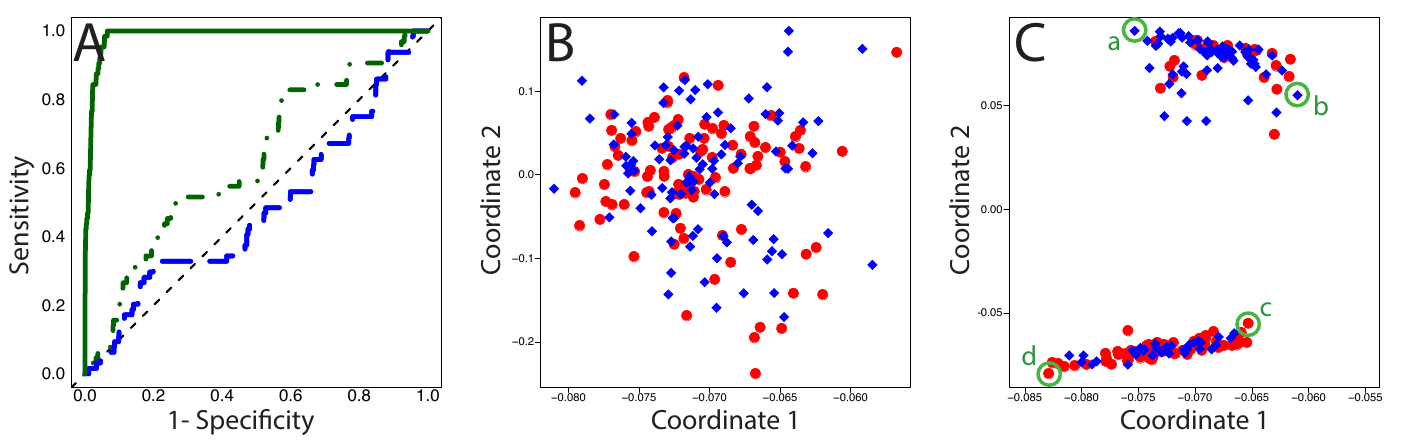}
\caption[Custom distance]{\textbf{A}: Mean ROC curves, over 8 separate RF iterations with $\gamma=0.3$ and $\delta=1.5$, for the detection of causal SNP pairs from brain connectivity, the latter represented by i) covariance matrices between the 400 brain ROIs, and ii) the SICE-inferred undirected graph representations. The performance of the distance-based RF regression is compared to the case-control RF classification RF run. The blue dashed curves represent the mean output from RF classifications with AD/CN case-control responses; the solid and dot-dashed green curves represent the mean output from RF regression using the graph distance \eqref{graphdist} and covariance metrics \eqref{covdist} respectively. \textbf{B/C}: Manifold plots deduced by applying Laplacian eigenmaps to the proximity matrices from RF classification and regression tasks performed using the covariance/graph metrics respectively. Each of the 200 shapes in each plot represents an individual from a simulated sample used in a given RF iteration. The \textbf{red circles/blue diamonds} colours provide the CN/AD labels of the subjects.}
\label{epimodel3}
\end{figure}

\begin{figure}[ht] 
\centering
\includegraphics{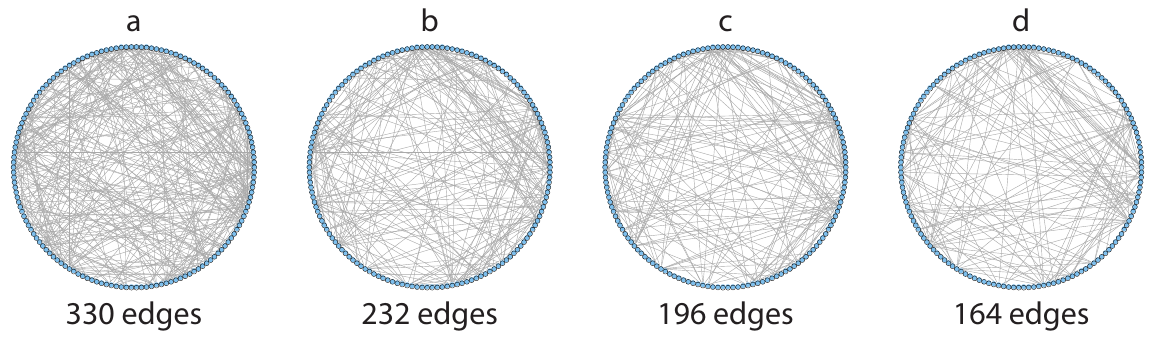}
\caption[SICE graphs]{SICE-inferred undirected graphs representing the brain connectivities of four subjects (points a-d from Figure \ref{epimodel3}, subplot C) of a single RFDM regression run using the graph metric \eqref{graphdist}. To avoid a cluttered graph plot, only a randomly selected 150 of the 400 ROI vertices are plotted here.}
\label{graphplots}
\end{figure}

\paragraph{Combined phenotypes}
In this experiment we combine, via the distance matrices $D_{ij}$, the volume gradients and the covariance matrix modalities for use in genetic association studies. To demonstrate the ability of RFDM to combine two relatively weak signals into a stronger amalgamated trait, we reduce the effect sizes for both quantitative phenotypes by halving the parameters, i.e. $\delta = 0.75$ and $\gamma = 0.3$.

We first perform two RF runs -- a RF classification with case-control responses and brain volume gradient covariates, and a distance-based RF regression on the genotype using a distance matrix derived from \eqref{covdist} (similar to Experiment 3). Taking the average proximity matrix \eqref{proxcomb} we derive the combined distance matrix, which we then use to perform a RF regression. The results are shown in Figure \ref{comb2}.

As a consequence of reducing the effect sizes, the causal SNP pair signals are relatively weak and the ROC curves for both quantitative responses now lie only slightly above the random diagonal. Nevertheless, by combining the proximity matrices we observe a significant boost in statistical power.

Besides the comparison of the shapes of the ROC curves (Figure \ref{comb2}), there is an alternative perspective on the process of combining modalities which involves the manifold plots. We examine the manifold plots arising from one (of the eight iterations) of the three RF analyses. In Figure \ref{manplots}, we see that, relative to the 2-dimensional manifold plots of the two individual modalities, the manifold plot from the RF regression utilising the averaged proximity matrix exhibits a pronounced clustering. We note also that due to the probabilistic disease classification model \eqref{bayespos}, this clustering, as one would expect, does not fully respect the case-control classification. However, the partitioning of the subjects according to the minor allele frequency of the simulated causal SNP with the largest marginal effect ($\text{maf} = 0\,\, \text{vs.} \,\, \text{maf} > 0$) exactly reproduces the two clusters. This demonstrates the ability of the combined modality distance matrix to capture the true underlying causal factors\footnote{We expect that by considering manifolds of dimensions $>2$, we will observe similar correspondences between the clustering and partitioning of weaker marginal SNP according to their maf.}. 

\begin{figure}[ht] 
\centering
\includegraphics{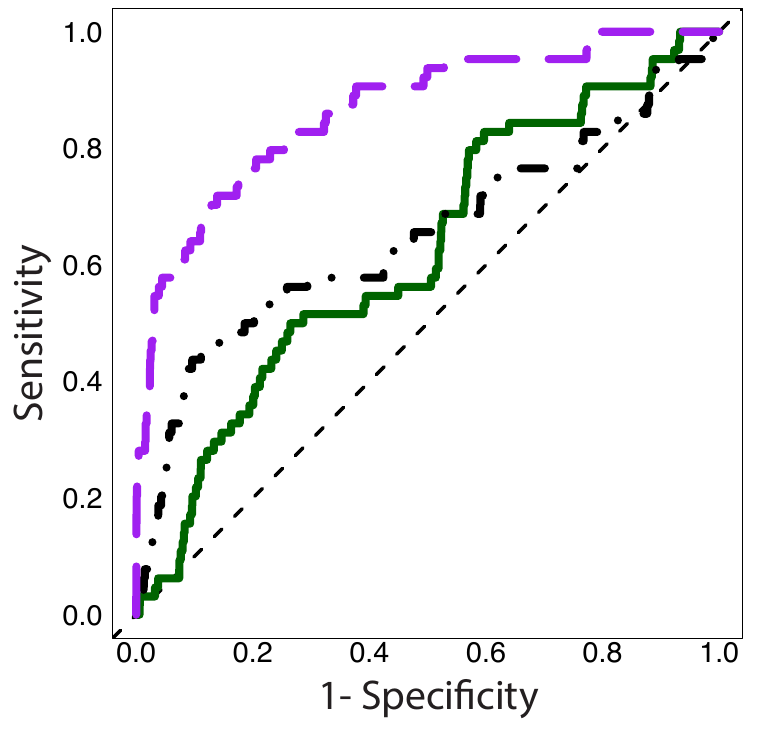}
\caption[Experiment 5: Combining modalities]{Mean ROC curves with $\gamma=0.3$ and $\delta=0.75$. The green solid curve represents the mean output from RF regression using the distance metric derived from the covariance matrix measure \eqref{covdist}; the black curve is the detection using a distance metric obtained through manifold learning of the imaging ROI variables from the case-control labels; the purple dashed curve represent the mean output from distance-based RF regression where the distance matrix is defined from the averaged proximity matrices from the other two RF runs (using covariance matrix distances (green curve) and manifold learning-obtained distances (black curve).}
\label{comb2}
\end{figure}

\begin{figure}[ht] 
\centering
\includegraphics{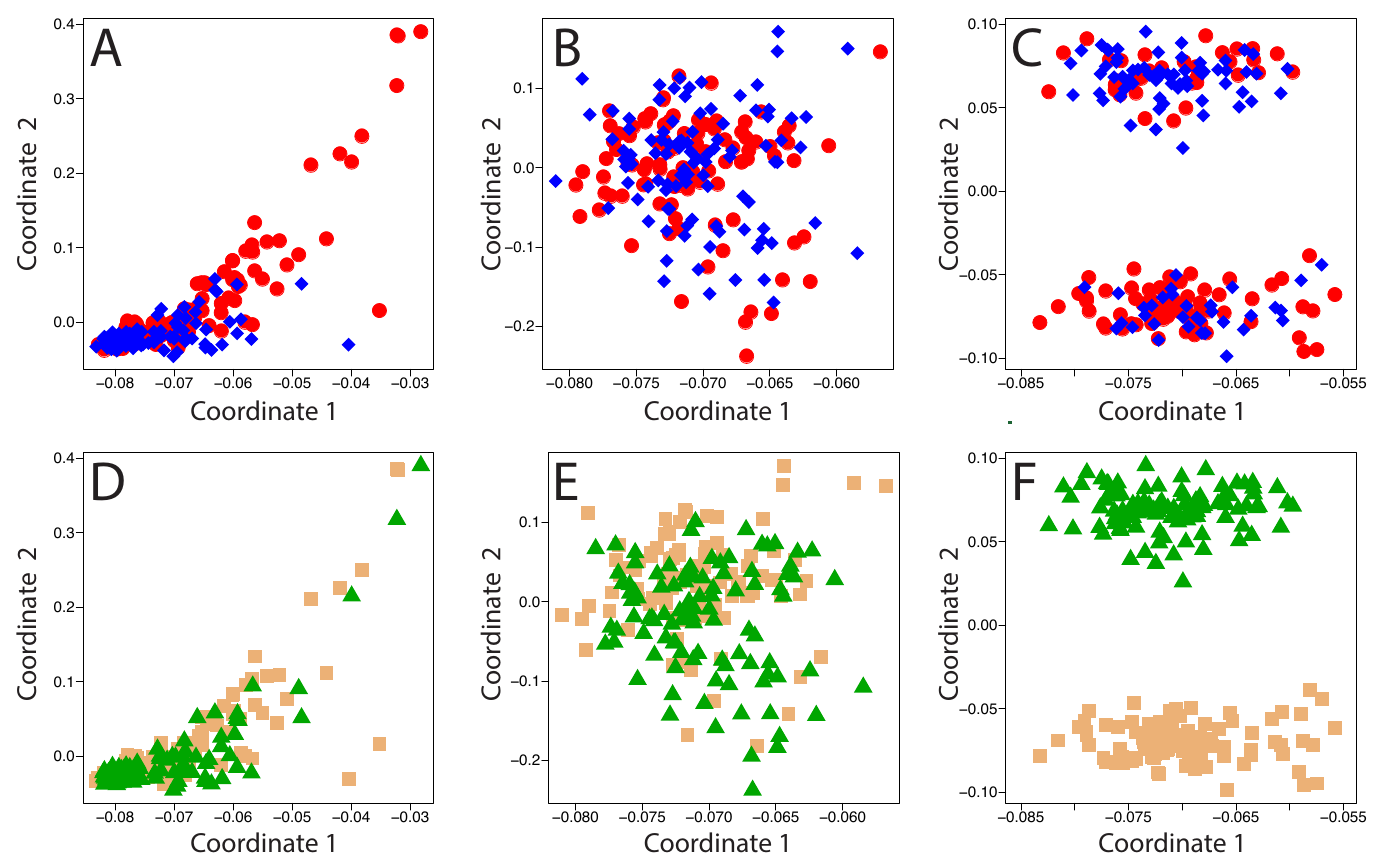}
\caption[Manifold learning plots]{Manifold plots deduced by applying Laplacian eigenmaps to the proximity matrices from RF classification and regression tasks involving the following modalities -- \textbf{A,D}: Brain ventricular volume gradient vectors (with case-control categorical response labels); \textbf{B,E}: Brain connectivity as represented by covariance matrices between the 400 ROIs (using \eqref{covdist} to define the distance matrix);  \textbf{C,F}: The combined quantitative trait (volume vector with covariance matrix) built from the distance matrix derived from the average of the two proximity matrices from the RF runs involving the individual modalities. Each of the 200 shapes in each plot represents an individual from a simulated sample used in a given RF iteration. The colours represent the following categorical labels of the different subjects. \textbf{red circles/blue diamonds}: control/cases; \textbf{green triangles/brown squares}:  $\text{maf} = 0\,\, / \,\, \text{maf} > 0$, where maf refers to the minor allele frequency of the simulated causal SNP with the largest marginal effect.}
\label{manplots}
\end{figure}

\paragraph{Spurious signals in the quantitative response vector}
This experiment explores the application of manifold learning methods in the construction of $D_{ij}$ and demonstrates its use in filtering out spurious signals in quantitative responses.

In addition to the set of potentially causal SNPs $S_c$, we now introduce a separate set of `spurious' SNPs $S_s$, whereby the corresponding effect on a subspace of the phenotypic vector space is similar to the causal effect. Specifically, we build another $P_3$ set by sampling from $S_s$ (instead of $S_c$) and implement the genetic effects model \eqref{volmod} on a non-overlapping subspace of $\mathsf{Y}$. This subspace is chosen to be twice the size of the disease-linked causal region $R_\text{d}$ (i.e. 46-dimensional). (See Figure \ref{model}). Because these genetic factors and their effect on the quantitative phenotype have no bearing on the propensity to develop the disease of interest\footnote{These additional effects may have possible links to other non-dementia related neurological pathologies. However, for the purposes of the detection of disease-linked SNPs or SNP-SNP pairs, these associations are irrelevant.}, they are classed as spurious effects.

It is obvious that the use of the Euclidean norm on the quantitative phenotypes is no longer appropriate; in cases where the strength of these spurious genetic factors is strong relative to the causal SNPs, the conclusions are possibly misleading. To demonstrate this we perform two sets analyses for similar ($=\delta$) and strong ($\delta \rightarrow 3\delta$) spurious effects sizes. As shown in Figure \ref{epimodel2}, for the weaker spurious effect, the use of quantitative phenotypes still presents added statistical power compared to using case-control labels. However in the case with a strong spurious effect, the causal SNP-SNP pair signals are not distinguishable, thereby resulting in a highly misleading variable-pair importance ranking (i.e. outperformed by a random selection).

We reduce the dimensionality of $\mathsf{Y}$, via manifold learning, to filter out the spurious signals. Using manifold forests, we perform the analysis in two stages. In the first stage, we perform a RF classification with the case-control classes as categorical responses and the quantitative imaging phenotype as the 400-dimensional covariate. From the resulting proximity matrix $W_{ij}$ \eqref{proximity} we obtain, via the use of Laplacian eigenmaps and by keeping only the first two manifold dimensions, the distance matrix $D_{ij}$. For the second stage, we perform a distance-based RF regression with the distance matrix $D_{ij}$ and the genotype as the covariate. The results are plotted in Figure \ref{epimodel2}. In both cases with weak and strong spurious effects, the distance-based RF method clearly gives the superior performance compared to both the case-control classification RF, and the multivariate RF regression with the naive implementation of the Euclidean norm.

\begin{figure}[ht] 
\centering
{\includegraphics{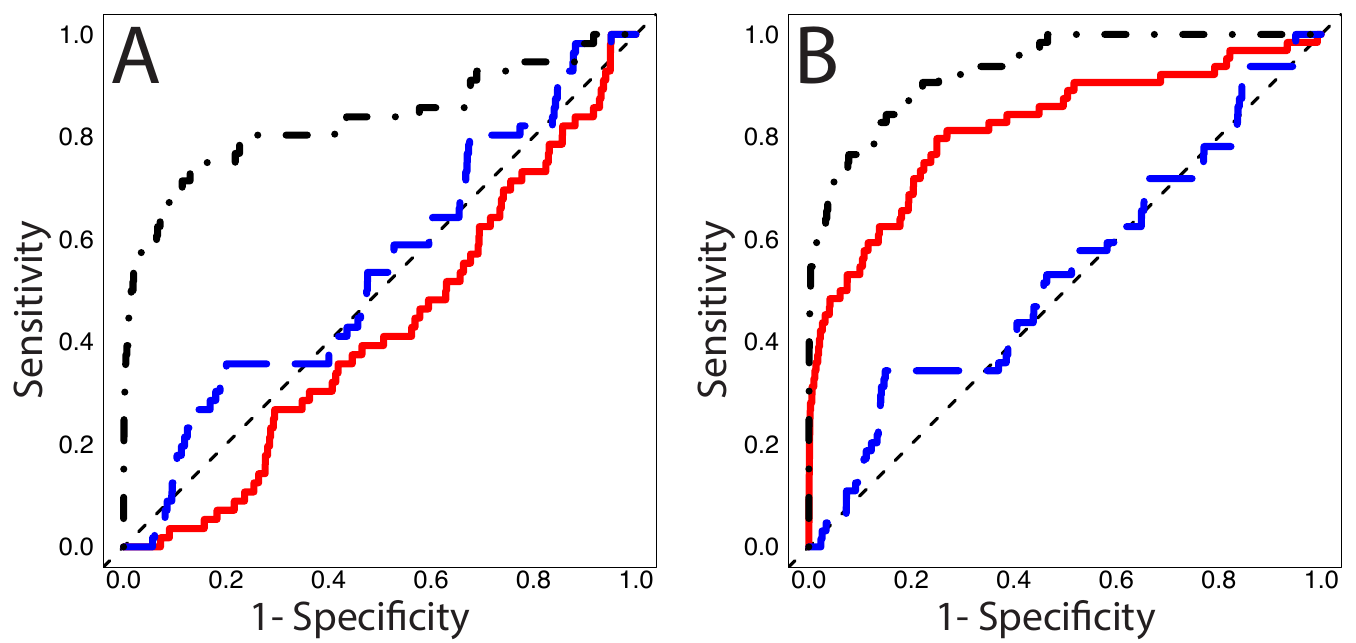}}
\caption[Spurious imaging]{Mean ROC curves for the detection of causal SNP-SNP pairs in the presence of spurious genetic effects, as averaged over 8 RF runs. \textbf{A}:  Strong spurious effect size ($\delta=4.5$); \textbf{B}: Weaker spurious effect size ($\delta=1.5$). The blue dashed curves represent the mean output from RF classifications with case-control responses; the red solid curves represent the mean output from RF regression with ventricular volume gradients as the multivariate responses (using Euclidean norm). The black dot-dashed curves represent the mean output from a distance-based RF regression, where the distance $D_{ij}$ is deduced from a manifold forest run using the the case-control labels as categorical responses and volume gradients as predictors.}
\label{epimodel2}
\end{figure}

\section{Application to Alzheimer's Disease} \label{realstudy}
We have applied our proposed methodology on real genotypic and phenotypic data extracted from the ADNI database. Our data is in the form of :
$$
\mathsf{S} = \bigl\{(\mathbf{x}_i,\mathbf{y}_i, \mathbf{z}_i)\,|\,\mathbf{x}_i\in \mathsf{X}, \,\mathbf{y}_i\in \mathsf{Y}, \,\mathbf{z}_i\in \mathsf{Z},\, i=1, \dotsc , N\bigr\}
$$
where $N=464$ is the total number of subjects, from which 99 exhibit AD, 211 are diagnosed with MCI and 154 are cognitive normal (CN) and serve as the controls in the ADNI study. For each subject, $\mathbf{z}_i \in \lbrace 0,1,2 \rbrace$ denotes the disease status, $\mathbf{y}_i$ is the $148,023$ dimensional vector describing the brain longitudinal slope coefficients, and $\mathbf{x}_i$ is the 7848 dimensional genotypic vector, with $x_{is} \in \{0,1,2\}$ being the minor allele count of each SNP feature (see Appendix \ref{datasets} ).

Due to the small sample size and in order to reduce the computational burden that whould have arisen with the inclusion of the whole genome SNP data (434,271 features per individual), we have restricted our analysis to genotypic data from chromosome 19, where two of the most prominent gene biomarkers for the detection of AD (namely APOE and TOMM40) are located.

The steps we took in our analysis are descibed below. First, we need to acquire a distance matrix indicative of the phenotypic variation between and within our sample population classes. Euclidean distance in the original high dimensional euclidean space cannot be considered a reliable metric. As such, we project the multivariate phenotypes into a low dimensional manifold, from which we then extract our distances. From the multitude of techniques that exist for dimensionality reduction, we chose to showcase two that incorporate the use of Random Forests, both in an unsupervised and a supervised setting. While most manifold learning techniques work in an unsupervised way, utilizing just the information from the $\mathbf{y}_i$ vectors, we aim to to enhance the quality of our manifold by incorporating the AD/MCI/Control classification $\mathbf{z}_i$ that is known for each endophenotype.  


For the case of unsupervised dimensionality reduction, we utilized the method of Totally Random Trees Embedding (TRTE), which is based on the decomposition of the endophenotypic covariance matrix, after applying a transformation of $\mathbf{y}_i$ to a high-dimensional binary representation \citep{Pedregosa2011,Moosmann2007,Busoniu2010}. Figure \ref{unsuper_man}(a) depicts the extracted manifold points for the 253 CN and AD subjects in our study, while in figure \ref{unsuper_man}(b) all 464 individuals are included. Euclidean distance was then used to extract the $464 \times 464$ pairwise distance matrix from the manifold.

\begin{figure}[ht] 
\centering
\scalebox{0.36}[0.36]{\includegraphics{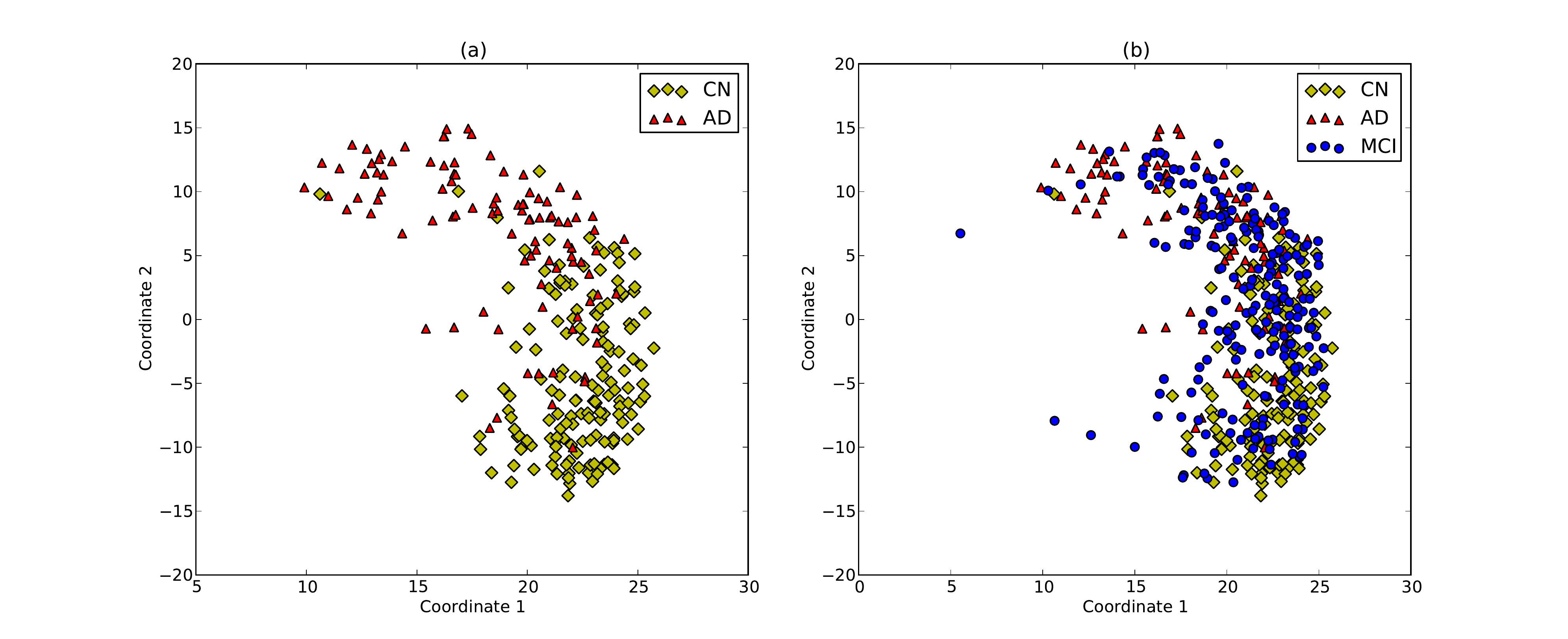}} 
\caption[2D manifold from Unsupervised method]{2D manifolds from Totally Random Trees Embedding \textbf{a}: with AD/CN subjects only; \textbf{b}: with all subjects.}
\label{unsuper_man}
\end{figure}

We also generate a phenotypic manifold by training a classification RF, with $\mathbf{z}_i$ as the response variables and $\mathbf{y}_i$ as predictors. By dropping the endophenotypic vectors in the trees we acquire a pairwise proximity matrix between all the subjects. Spectral decompotition of that matrix provides the $\mathbb{R}^2$ manifold representation of our phenotypic data. Figures \ref{super_man}(a) and \ref{super_man}(b) depict this manifold. 2 dimensions were deemed sufficient to capture the distribution of the data for our purposes, based on the eigenvalues of the matrix. We can notice that in the second manifold we get a reduced variance within each class, especially for the AD and Control subjects, compared to the unsupervised manifold. 


\begin{figure}[ht] 
\centering
\scalebox{0.36}[0.36]{\includegraphics{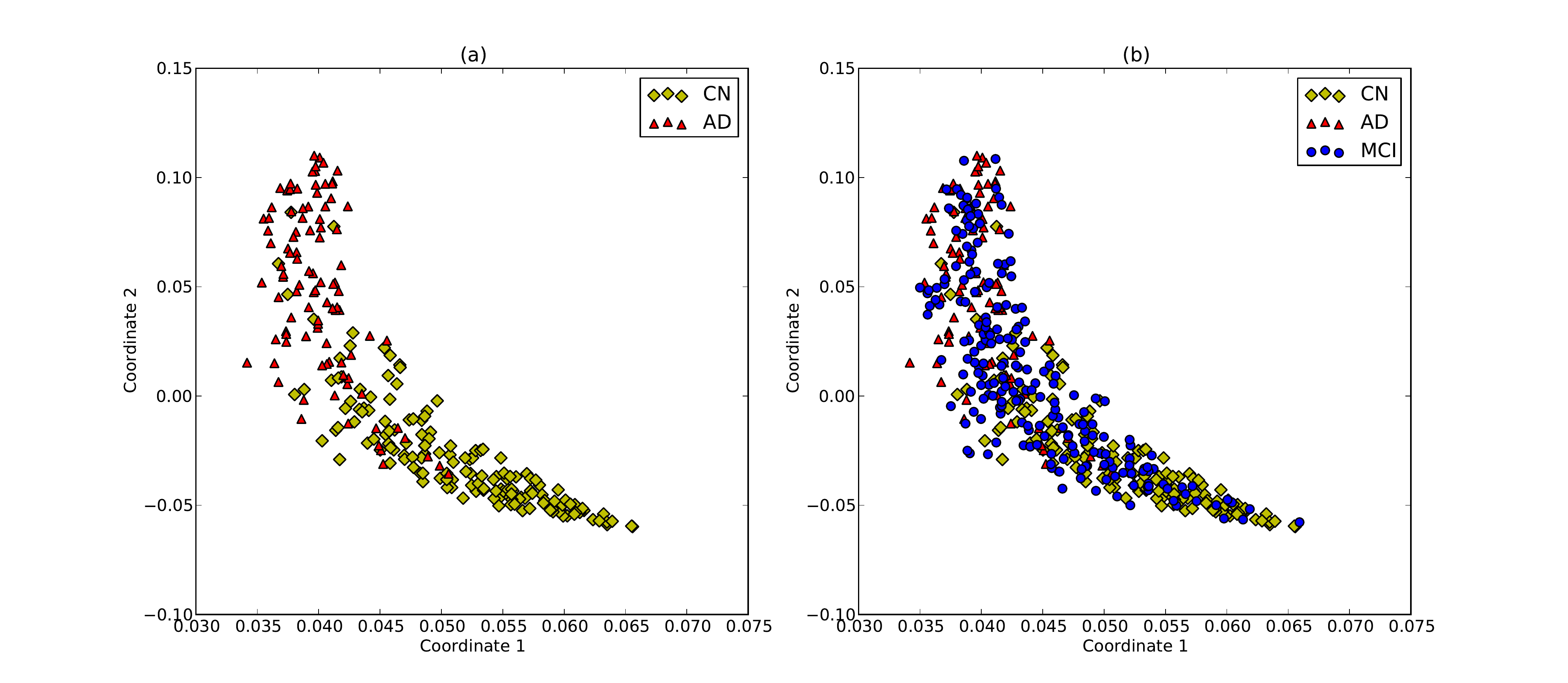}} 
\caption[2D manifold from Classification Random Forest]{2D manifolds from Classification Random Forest \textbf{a}: with AD/CN subjects only; \textbf{b}: with all subjects.}
\label{super_man}
\end{figure}


Using the distance matrices to describe the phenotypic responses and the multivariate genotypic data as input features, we train two RFDM using pairwise distances from the unsupervised setting and two utilizing distances from the supervised one. In order to validate the ability of our method to extract well known marginal and epistasis results, we first use the whole set of 7848 SNP features. Consequently, we exclude the APOE$\varepsilon$4 SNP and rs157580, rs2075650, rs8106922 SNPs -- belonging to TOMM40 gene -- from the genotypic features, in order to reveal epistatic interactions that were masked previously by the large marginal effects of the known AD associated SNPs.

Each forest consisted of 4000 trees, with a minimum node size of 5 subjects and maximum tree depth equal to 6. The training and analysis was performed on a 15-node private cluster, where each node is equipped with 24GB memory and 6 Intel(R) Xeon(R) 2.27GHz CPUs. 

Table \ref{marg_table} presents the SNPs with the highest information gain scores, providing a measure of top marginal effects. It is clear that known SNP biomarkers for AD appear in the top places of the runs with all chromosome 19 SNPs. The fact that the most significant associated genes on chromosome 19 are top ranked in our search provide additional evidence for the effectiveness of RFDM in GWA studies. Furthermore, a literature search reveals that other genes appearing on the lists, namely INSR, GPI, PVRL2 and BCL3 have been previously connected to Alzheimer's disease \citep{PERICAKVANCE1991,IWANGOFF1980,Zhao2009}. By plotting the information gain scores of the ranked results (plots not shown here) we notice an expected behaviour of values rapidly decreasing after the top 2 SNPs, for the studies with all biomarkers, and after the top 7 SNPs, when we exclude the aforementioned four SNPs. This increase in the number of distinguished rankings is justifiable by the fact that these SNPs have strong marginal effects that mask the lesser but still existing infuence of other features.

\begin{table}[ht]
\small
\centering
\caption{Top marginal scoring SNPs and corresponding genes from 4 RFDMs. Entries marked with an asterisk (*) have known associations with Alzheimer's disease (source: http://www.ncbi.nlm.nih.gov/snp). \textbf{(a)}: all SNPs, with distance matrix extracted by unsupervised method; \textbf{(b)}: excluding APOE and TOMM40 SNPs, with distance matrix extracted by unsupervised method; \textbf{(c)}: all SNPs, with distance matrix extracted by classification RF; \textbf{(d)}: excluding APOE and TOMM40 SNPs, with distance matrix extracted by classification RF.}
\resizebox{\textwidth}{!}{
\begin{tabular}{l l | l l | l l | l l}
\hline
\hline 
\multicolumn{8}{c}{Marginal Effects}
\\
\multicolumn{2}{c}{(a)} & \multicolumn{2}{c}{(b)} & \multicolumn{2}{c}{(c)} & \multicolumn{2}{c}{(d)}
\\
\hline
APOE$\varepsilon$4 $^*$ & (APOE) & rs439401 $^*$ & (-)  & APOE$\varepsilon$4 $^*$ & (APOE) & rs439401 $^*$ & (-)
\\
rs2075650 $^*$ & (TOMM40) & rs889130 & (-) & rs2075650 $^*$ & (TOMM40) & rs892023 & (-)
\\
rs439401 $^*$ & (-) & rs11666282 & (-) & rs157580 $^*$ & (TOMM40) & rs889130 & (-)
\\
rs157580 $^*$ & (TOMM40) & rs4802839 & (-) & rs4806015 & (GPI) & rs10415849 &  (-)
\\
rs889130 & (-) & rs892023 & (-) & rs12608634 & (-) & s440277 & (PVRL2)
\\
rs4802839 & (-) & rs10415849 & (-) & rs889130 & (-) & rs8103315 & (BCL3)
\\
rs8103483 & (INSR) & rs4803150 & (-) & rs439401 $^*$ & (-) & rs1103851 & (ANKRD27) 
\\
rs4806015 & (GPI) & rs2277968 & (COL5A3) & rs4344866 & (-) & rs12608634 & (-)
\\
rs10415849 & (-) & rs8103483 & (INSR) & rs892023 & (-) & s8105340 & (PVRL2)
\\
rs4344866 & (-) & rs10406152 & (COL5A3) & rs440277 & (PVRL2) & rs918484 & (TMIGD2)
\\
\hline
\end{tabular}}
\label{marg_table}
\end{table}

Table \ref{epis_table} shows the two lists with the top 10 ranked interaction SNP pairs, from the runs without APOE and TOMM40 SNPs. It is evident that the variability of results is increased here, which is expected, taken into account the combinatorial nature of the process, which means $7846\times(7846/2) -7846 = 30772012$ ranked feature pairs for each run. There are though noticeable agreements between the top results of the lists.

\begin{table}[ht]
\small
\centering
\caption{Top interaction scoring SNP pairs and their corresponding genes from 2 RFDMs excluding feature SNPs APOE$\varepsilon$4, rs157580, rs2075650 and rs8106922 . \textbf{(a)}: From RFDM with distance matrix extracted by unsupervised method; \textbf{(b)}:  From RFDM with distance matrix extracted by classification RF.}
\resizebox{\textwidth}{!}{%
\begin{tabular}{l l | l l}
\hline
\hline 
\multicolumn{4}{c}{Epistatic Effects}
\\
\multicolumn{2}{c}{(a)} & \multicolumn{2}{c}{(b)}
\\
\hline
rs10415849-rs889130 & (-)-(-) & rs439401-rs606119  & (-)-(AZU1)
\\
rs440277-rs889130 & (PVRL2)-(-) & rs440277-rs889130 & (PVRL2)-(-)
\\ 
rs4802839-rs11666282 & (-)-(-) & rs439401-rs12460890 & (-)-(AZU1)
\\
rs439401-rs606119 & (-)-(AZU1) & rs10415849-rs889130 & (-)-(-)
\\
rs439401-rs889130 & (-)-(-) & rs440277-rs8182496 & (PVRL2)-(-)
\\
rs439401-rs2277968 & (-)-(COL5A3) & rs11666426-rs439401 & (ZNF665)-(-)
\\
rs4803150-rs439401 & (-)-(-) & rs439401-rs440277 & (-)-(PVRL2)
\\
rs439401-rs10406152 & (-)-(COL5A3) & rs892023-rs2240747 & (-)-(-)
\\
rs10415849-rs10406152 & (-)-(COL5A3) & rs439401-rs405509 & (-)-(-)
\\
rs1725474-rs11672955 & (SIPA1L3)-(RAX2) & rs439401-rs12460918 & (-)-(-)
\\
\hline
\end{tabular}}
\label{epis_table}
\end{table}

Detailed evaluation of the interaction SNP pairs is outside the scope of this article, since it would require elaborate analysis and exhaustive biological annotation. By mapping the different SNPs in our epistasis lists to their genes we are able to exploit gene-expression data and provide a short assessment of our results. Due to the fact that most SNP pairs do not have corresponding genes, we had to go further down the interaction lists to identify associations between known genes. Three out of the eight top gene interacting pairs we extracted, namely ZNF160 - INSR, PVRL2 - TJP3 and SIPA1L3 - RAX2 have supporting evidence for their co-expression profiles, from microarray studies related to neurological diseases \citep{Warde-Farley2010}. In more detail, each of the aformentioned pairs were reported of having significant pearson correlation values in respective microarray studies of bipolar disorder and schizophrenia \citep{Iwamoto2005}, parkinson's disease \cite{Scherzer2007}and autism \citep{Alter2011}.

\section{Summary and discussion}\label{summ}

In this paper we have proposed Random Forests on Distance Matrices as a generalisation of the RF regression machine learning tool for imaging genetics studies. The key feature of this novel methodology is the use of distance matrices in place of the usual univariate or multivariate quantitative responses. 

As outlined is this paper, this generalisation to a representation-independent approach has several advantages, not the least one being to open up the possibility of analysing non-vectorial data as one would for standard Euclidean vectors. We propose a general framework in which to perform regression on any data type for which there exists either a definition of or a method to learn the distances between data subjects. Although not strictly part of the RFDM algorithm, the process of learning these distances can be inferred from the RF proximities, the latter obtained via a separate RF run. We showed that employing this supervised distance learning step can be effective in filtering out the spurious signals from our dataset.

The representation-independent approach has coincidentally allowed us to combine datasets of (possibly) different representations by averaging over distances matrices. The combination of datasets in the context of RF proximity matrices has previously been utilised in \citep{gray}; however the purpose was the improvement of RF predictions as opposed to identification of variable importances.

We have validated the methodology by means of a semi-empirical Alzheimer's disease genetic model in an imaging genetics simulation framework. Utilising longitudinal MRI ventricular volume data from ADNI and HapMap phase 3 data to simulate a population brain atrophy pattern and genetic profile respectively, we constructed a semi-realistic AD model and implemented a wide range of population penetrance and effect size assumptions. In addition, we also simulated basal and the AD-attenuated brain connectivity graphs. We illustrate over several simulation experiments the above advantages of RFDM methodology. 

Promisingly, the indications are that RFDM is particularly suited for the utilisation of brain connectivity data, inferred from fMRI scans \citep{brier} or PET data \citep{sun}, for genetic association studies. In studies on brain connectivity changes in AD, the focus is typically on the gaining insight into the pathology of AD; so far, to the authors knowledge, RFDM is the only method which will accommodate brain network data in imaging genetics studies. 

Combining datasets is perhaps the largest immediate benefit of the methodology in imaging genetics. In addition to using connectivity data, one can also combine the various defined or inferred distance matrices from different modalities. In our simulations, we showed that combining two datasets with (deliberately designed) weak effects significantly boosted the overall signal strength of the causal SNPs. This capability is expected to be useful for AD where, apart from a handful of strongly-linked AD-associated genes, the majority of genetic factors are expected to only have weak marginal effects.

Widening the scope beyond AD to more general diseases where individual-gene effects are strong, RFDM, and RF in general, are still crucial in the search for the genetic disease factors. The driving factor in this is the recognition of the increasing need to understand non-linear effects and interactions between genes. As widely recognised in the literature, due to its hierarchical branching tree structure, RF are particularly suited to probe for epistasis effects. However, it is often claimed (see, for example, \citep{geneRF}) that although RF variable importances factor in non-linearity, the algorithm is unable to identify specific interacting pairs. We have overcome this limitation via our proposed Gini pairwise interaction measure which ranks the importance of two-way interacting variable pairs, just as one would rank the original, marginal, variables. Through a range of interacting models, we show in our AD simulation experiments, that the conditional Gini index is capable of identifying epistasis. However we note that, by construction, the conditional Gini index does not possess equitability; the detection of interacting pairs is dependent on the marginal importance of the individual variables themselves, where pairs with one or two strong marginal importances will rank higher than a pair with similar interacting strength but with weaker marginal importances. Therefore, one should avoid interpreting the conditional Gini index as a \textit{measure} of the strength of the interaction.

One advantage of using RF to search for epistasis, compared to other existing methods, is that the calculation of the conditional Gini information uses only the original Gini indices and, therefore, the additional computation effort is insignificant. In principle this process can be extended to higher-order interactions. For example, a three way interaction can be quantified by the conditional Gini term
\begin{equation}\nonumber
G_{\alpha\beta\gamma} \equiv G_{\alpha|\beta\gamma} + G_{\beta|\alpha\gamma} + G_{\gamma|\alpha\beta},
\end{equation}
where the individual terms are defined in analogy with the pairwise expressions \eqref{ginipairindex}. For such higher-order terms, one must satisfy two conditions for the interaction scores to be precise. Firstly, the number of trees in the RF must be high enough to ensure that each three-feature combination appears a reasonable number of times. Secondly, the dataset must be large enough to ensure that the depths of the trees are sufficiently deep to accommodate the number of features. It is for these two reasons that the use of the conditional Gini importance measure is less suited for the detection of higher-order interactions. Nevertheless, one can infer the possible presence of a higher-order interactions by detecting communities in the pairwise interaction network.

As for RFDM, there is the added benefit that using a distance matrix approach for standard multivariate vectorial data incurs a relatively cheap one-off computational cost at the start of the run for a measurable reduction in total computational time. The reason is that calculating squares of matrices is relatively quicker than taking the sum-of-squared differences to the mean.

The detection of SNPs and interacting SNP-SNP pairs has been the focus of this work. One drawback of RFDM is in the original use of RF regression, namely as a prediction tool for out-of-sample data. Passing an external data through a completed RF will require knowing the distance between each new subject and the subjects in the training data set, the latter being represented by the distance covariance matrix. If a reference distance measure was used, this is not a problem as the distance matrix can simply be augmented by a row and column representing the given additional subject; when the original distance matrix is learned rather than defined, this simple augmentation is no longer possible. Instead, one projects each out-of-sample data point onto the manifold plot and calculate the distances using the usual Euclidean norm. However, should this datapoint differ significantly from the training data set subjects, this projection will not be as good an approximation to the actual distance as that obtained by training the RF with the additional subject in the test set. For imaging genetics, fortunately, this downside is moot as one is primarily interested in variable importances.

Although the above simulations and real studies are done in the context neuroimaging genetics, both the simulation framework itself and our particular version of the RF algorithm are completely general and we expect the overall conclusions, if not the results, to be applicable to other multivariate regression problems.

There are several research directions to take. The most immediate task is to conduct a GWAS utilising single and combined modalities using RFDM. In this paper we have endeavoured to simulate a realistic model to validate and test the power of the methodology.  Although we have chosen a smaller imaging and genomic datasets for the sake of being able to perform multiple (computationally expensive) RF run across several scenarios and assumptions, preliminary studies on much larger datasets indicate that the method remains valid for higher-dimensional datasets. Also, the modelling of brain connectivity assumed a monotonic decline in brain connectivity. However, recent studies have shown that certain connectivity between ROI initially increases in the early stages of AD progression, which the authors speculate is a compensatory effect in response to loss of connectivity elsewhere in the brain. One would like to model this effect and test different graph distance measures in this context.

A Python implementation of the Distance RF methodology, and the Python and R scripts for the data simulation procedure is available upon request. 

\section*{Appendix}

\subsection*{Data sets} \label{datasets}


Concerning the simulation studies, for the purposes of realism, the simulated genotypes and quantitative phenotypes are built from the public datasets of the International HapMap Project\footnote{http://hapmap.ncbi.nlm.nih.gov/} and the Alzheimer’s Disease Neuroimaging Initiative (ADNI)\footnote{http://adni.loni.ucla.edu/data-samples/} respectively. The genotypic and phenotypic data that is used for the application on Alzheimer's disease was also retreived from ADNI.

\subsubsection*{HapMap data}
A human haplotype map of over 1.5 million SNPs for 993 subjects from 11 sub-populations across four continents was obtained from the second release of the HapMap phase 3 dataset\footnote{ftp://ftp.ncbi.nlm.nih.gov/hapmap//phasing/2009-02\_phaseIII/HapMap3\_r2/} of the International HapMap Project.

\subsubsection*{ADNI data}
The ADNI study began at 2004 as a multicenter effort to develop clinical, imaging, genetic and biochemical biomarkers for the early detection and tracking of AD. The initial six-year phase of the study (ADNI1) included 400 subjects diagnosed with mild cognitive impairment (MCI), 200 subjects with early AD and 200 elderly control subjects. In 2009, ADNI1 was extended with ADNI GO which assessed the existing ADNI1 cohort and added 200 participants identified as having early mild cognitive impairment (EMCI). Finally, in 2011, ADNI2 began, which assesses new participants in addition to those from the ADNI1/ADNI GO cohort. All participants in the ADNI studies were recruited across North America and are followed and reassessed over time to track the pathology of the disease as it progresses.


For the purpose of applying distance-based Random Forests on a real dataset, we acquired genotypes for 464 subjects from the ADNI database ((99 AD, 154 CN, 211 MCI). Genotyping was performed with the Human610-Quad Bead-Chip microarray, which includes 620,901 SNPs and copy number variations (see \citep{Saykin2010} for details). A separate genotyping of the SNPs for the APOE$\varepsilon$4 variant has been performed by ADNI, since these were not included in the above microarray. The subjects were unrelated, of European ancestry, and passed screening for evidence of population stratification using the procedure described in \citep{mulm}. 

The selection and preprocessing steps are taken from \citep{silver}, where the same data was used for the identification of possible AD causal gene pathways.  In detail, the genotypic data consists from autosomal SNPs with genotyping rate $>95\%$ (42,680 SNPs), Hardy–Weinberg equilibrium p-value $>5\times 10^{−7}$ and minor allele frequency $>0.1$. For each subject we ended up with a set of 434,271 SNPs. Any missing genotypes were imputed using the same procedure as in \citep{vounou2}.


The imaging data consists of longitudinal brain MRI scans (1.5 T) from the ADNI database. For each of the subjects, serial brain MR images were acquired at baseline and at 6, 12 and 24 months following initial recruitment. Details about the acquisition protocol and the preprocessing steps that were applied by ADNI on the raw scans can be found in \citep{Jack2008}.

Further processing was carried out, as described in detail in \citep{silver}. The aim is to provide a measure of brain structural change over time. To achieve that, a linear regression model with an intercept term and time as the independent variable is fitted for each voxel. The coefficient for the slope provides a measure of brain tissue change, relative to time, for each voxel. Correction for age and sex, as well as selection of the most discriminative voxels is performed by analysis of variance (ANOVA) \citep{silver}. The final endophenotype data that we use is in the form of longitudinal maps of ventricular/CSF expansion (reconstructed to 1mm isotropic voxels), which represent brain tissue loss and are normalised across all the subjects. Each subject's endophenotype vector consists of 148,023 slope coefficients.

The simulation studies below were performed using information from the subset of CN and AD subjects, while for the application study, MCI subjects were included as well.


\subsection*{Imaging genetics simulation engine}\label{simdetails}



In this section we provide details of the imaging genetics simulation engine. We describe, in order, the genotype simulation, the phenotype simulation, the genetic disease modelling, and lastly the disease classification modelling.


\subsubsection*{Base genotype simulation}

The ADNI dataset with 253 subjects is clearly too small to represent a direct population source for the multiple samples required for the multiple simulation runs. Instead we simulate this base population and its genotype within an individual-based, forward-time, population genetics simulation environment. Specifically we employed the \textit{SimuPOP} Python package \citep{simupop} and have made use of the \textit{SimuGWAS}\footnote{http://simupop.sourceforge.net/cookbook/pmwiki.php/Cookbook/SimuGWAS} implementation \citep{peng}, originally written for the simulation of samples for GWAS studies. This allows one to evolve the genotype of a given starting population over a specified number of generations with exposure to the standard evolutionary forces of mutation, recombination, migration, and selection.  

The pattern of linkage disequilibrium (LD) between markers in a forward-time simulated population is sensitive to the make-up of the starting population. In the absence of data on populations in the distant past, we adopt the HapMap phase 3 subjects as a reasonable proxy. Rather than evolving the entire genotype, it is suffices for our simulation study to restrict our attention to a subset of genetic markers located on a single chromosome. From all 993 subjects in the HapMap dataset, we extract 385 sequential, biallelic, SNP markers from the front end of chromosome 19 with a minimum MAF of 0.05. We perform a forward-time evolution of this initial population of 993 individuals over 500 non-overlapping generations using the Wright-Fisher model. We assume a constant population growth rate towards a final population of $10000$ individuals. We maintain a recombination rate of $1.0 \times 10^{-8}$, a mutation rate of $1.0 \times 10^{-8}$, and a migration rate of $1.0 \times 10^{-3}$ throughout the simulated evolutionary process.

We denote the genotype of individual $i$ at locus $s$ by the minor allele count $x_{is} \in \{0,1,2\}$. From the set of 385 markers, we identify two subsets of 16 SNPs with $\text{MAF} \sim 0.2$\footnote{The specific ranges selected are $0.195 < \text{MAF} < 0.205$ and $0.22 < \text{MAF} < 0.24$ respectively. The choice of 0.2 was made solely to identify loci with a clear difference between their major and minor allele frequency and is otherwise arbitrary.}, the purpose of which will be made explicit later in Section \ref{dismod}. Let the two subsets be $S_{\text{c}}$ and $S_{\text{s}}$, where the subscripts indicate the labels \textit{potentially causative} and \textit{potentially spurious} respectively. We emphasise the potentiality in their descriptions to highlight the fact that, depending on the specific genetic model simulated (see later), not all the SNPs from the two sets will be associated to phenotypic modifications.

We consider a single chromosome with 385 markers and denote the genotype of individual $i$ at locus $s$ by the minor allele count $x_{is} \in \{0,1,2\}$. We assign two subsets of SNPs, $S_{\text{c}}$ and $S_{\text{s}}$, which we label \textit{potentially causative} and \textit{potentially spurious} respectively. We emphasise the potentiality in their descriptions to highlight the fact that, depending on the specific genetic model simulated (see later), not all the SNPs from the two sets will be associated to phenotypic modifications.

\subsubsection*{Base phenotype simulation}\label{endosim}
There is no constraint on the number or type of imaging phenotypes for the simulation model. In this paper we consider two such quantitative phenotypes -- a multivariate, vectorial, representation and a matrix representation of imaging phenotypes. These representations are motivated by both the neuroimaging data currently available, and also by the pathology of AD. For instance, the neuronal atrophy in AD patients can be represented as vector of time-averaged rates of volume change of the various ROIs in the brain.  Also, as demonstrated in several more recent studies \citep{sun, brier, huang}, the number of connections between different ROI is shown to be lower in AD subjects as compared with CN subjects. A similar decrease in connectivity is also observed in longitudinal FDG-PET and resting-state functional MRI (rs-fcMRI) scans of individual AD subjects \citep{sun, brier}. Although there are exceptions at the beginning of AD progression where connections between certain brain regions increase, it is believed that such localised changes are simply compensatory responses to the overall decline in brain connectivity. As the disease progresses, the exceptional trends reverts to the global pattern of a decrease in connectivity. Therefore, for the second endophenotype, we consider a matrix-valued phenotype representing brain-networks.


We denote the base vectorial phenotype of the $k$th ROI of individual $i$ by $y_{ik}$. Although we generate these vectors from the ADNI data representing voxel-wise ventricular volume expansion rates, the simulation framework is not dependent on the chosen method; indeed any suitable method for generating realistic vectors representing alternative imaging phenotypes is entirely permissible.


We plot the linear regression slope coefficient of the ventricular volume against time and identify, for the 253 ADNI subjects, a $148023$-voxel subset of the full-brain MRI scan which is maximally discriminative between AD and CN. Given that we aim to perform multiple RF runs over multiple scenarios, we perform a simple $k$-means clustering, transforming the $148023$-voxel ventricular volume gradients to that of just 400 parcels, simulating a lower-resolution image. This choice is arbitrary except for the fact that it is larger than the sample size, thereby simulating a large-$p$/small-$N$ scenario.

For each individual in our simulated population, the aim is simply to generate a 400-dimensional vectorial phenotype to represent the base ventricular gradients of the 400 ROIs on which the genetic model modifications can be imputed. We require, for realism, that \textit{after} implementing the genetic modifications, the multivariate distribution of this phenotype resembles the empirical distribution of the subjects from the ADNI dataset. The practical implementation of the base ventricular volumes is performed in two steps, as follows. 

For the first step, we generate random samples from a multivariate Gaussian distribution where the mean and covariance matrix parameters are taken from the sample mean and sample covariance matrix of the 154 CN subjects. Note that because there are fewer subjects than ROIs ($154 < 400$), the sample covariance matrix is unstable, requiring, therefore, the use of a James-Stein-type shrinkage estimator \citep{strimmer} for the covariance matrix. The distribution of rate gradients across all parcels of cognitive normal (CN) subjects has an average and standard deviation of -0.38 and 2.27 respectively, reflecting an overall decrease and variability in ventricular volume gradients in the base population across all the parcels. We denote the imaging trait of the $k$th ROI of individual $i$ by $y_{ik}$. 

For the second step, in anticipation of our genetic effects model where an overall \textit{increase} in each vector component is observed, we modify the above random samples by \textit{reducing} the gradient for selected ROIs. As shown in \citep{silver}, the brain matter atrophy is not uniform throughout the brain. We simulate this effect by selecting a subset of $n_d=28$ randomly selected ROIs which we label \textit{disease-affected}-ROIs, which corresponds to regions affected by selected SNPs which, in turn, has a measurable on the disease status. We label the ROI-index subset by $R_{\text{d}}$. Therefore the modification in this second step is 
\begin{equation}\label{tuning}
y_{ik} \quad\longrightarrow\quad y_{ik} - \zeta, \quad \forall k\in R_{\text{d}},
\end{equation}
where $\zeta >0$ is a parameter which is tuned according to the disease model (see below).



It is not unreasonable to assume that the volumes, and by extension the rate of volumetric changes, of functionally interconnected brain regions correlate positively \citep{autism}. This link between structural and functional aspects can be used to indirectly estimate brain connectivity networks via the atrophy gradients. We simulate the brain connectivity by first considering the full (non-partial) correlations between the same 400 ROI. From the $n_{CN}=154$ subjects, we have the empirical covariance matrix
\begin{equation}
S_\Sigma = \frac{1}{n_{CN}}\sum_{i=1}^{n_{CN}}(\mathbf{y}_i - \mu_\mathbf{y})(\mathbf{y}_i - \mu_\mathbf{y})^T.
\end{equation}
We generate the base covariance matrices for each of our 200 subjects by introducing random perturbations from the unit Normal at each component
\begin{equation}\label{basecov}
\Sigma^{(i)}_{kl}= (S_\Sigma)_{kl} + s^{(i)}_{kl},
\end{equation}
with $s^{(i)}_{kl} \sim \mathsf{N}(0,1)$ for subject $i$. We symmetrise the resulting matrix $\Sigma^{(i)}_{kl} \longrightarrow \frac{1}{2}(\Sigma^{(i)}_{kl} + \Sigma^{(i)}_{kl})$, ensuring that it remains positive definite \citep{higham}.

The graph representation of the network is inferred by identically matching the zero-components of the adjacency matrix of $G_i$ with the zeros of the the sparse inverse covariance estimate (SICE) of $\Sigma^{(i)}_{kl}$ \citep{sun, huang}. Let $\Theta^{(i)} = (\Sigma^{(i)})^{-1}$ be the inverse covariance matrix. Assuming that the observations follow a multivariate Gaussian, the SICE is obtained by maximising over $\Theta$ the $L_1$-penalised log-likelihood, 
\begin{equation}\label{SICEmax}
\log \det \Theta - \tr(S_\Sigma\Theta) - \rho ||\Theta ||_1,
\end{equation}
where $||\Theta||_1$ is the sum over the absolute values of all elements in $\Theta$. We use the graphical lasso algorithm \citep{glasso} to estimate $\Sigma$, and hence $\Theta$. The regularisation parameter is set as $\rho=1$ throughout.

One example of such a simulated brain network in the context on AD is found in \citep{sun} where the authors used the SICE approach on FDG-PET data to uncover connectivity among different brain regions.

The brain connectivity network of each subject is represented by a 400-dimensional square matrix, representing either the covariance between the different ROIs or the adjacency matrix of a 400-node graph. In this paper we consider both cases. The covariance and adjacency matrices are written as $\Sigma^{(i)}_{kl}$ and $G_i$ respectively.


\subsubsection*{Genetic disease modelling}\label{dismod}

The second step in our simulations is the construction of the disease models. There are two aspects to this -- the genetic effects on the endophenotypes, which includes both disease-related and `spurious' non-disease related effects, and the disease classification model.


Let $y_{\beta}$ be the simulated baseline vectorial phenotype at the ROI with index $\beta$. We assume a simple additive genetic model where the components of the vector modified from the base value are given by
\begin{equation}
y^*_\beta = y_\beta + w_\beta,
\end{equation}
where the effect size $w_\beta$ is built from multiplicative interaction effects as
\begin{equation}\label{volmod}
w_\beta = \delta_\beta\Bigl(\sum_{(a,b)\in P_\cdot}x_ax_b+ \sum_{c\in P_\cdot} x_c\Bigr).
\end{equation}
$P_\cdot$ refers to the interaction causal models above, and $x_a=\{0,1,2\}$ is the minor allele frequency at locus $a$, etc.
$\delta_\beta$ is an effect-size parameter which, when increased, with all else equal, monotonically boosts the population disease penetrance levels (see \textit{Disease classification models} below). For simplicity, we fix $\delta_\beta = \delta$ for all affected ROI.

In selected simulation runs, we include the presence of `spurious' SNP pairs and brain ROI (see Figure \ref{model}). These are labelled `spurious' because the genetic effects on the brain phenotype do not affect the propensity to develop the disease of interest. In our simulations, we assume that these do not overlap with their `causal' counterparts. Using the identical tuning parameter $\delta$ we employ a similar genetic effects model as above.

For the changes to brain networks, we propose an additive genetic model, similar to \eqref{volmod} where the strength of correlations between different ROI is scaled with a simple exponential factor. Let $\Sigma'_{kl}$ be the covariance matrix representing the brain connectivity. The components of the attenuated representative covariance matrix $\Sigma'^*$ are given by
\begin{equation}\label{covmodel}
\Sigma'^*_{kl} = \Sigma'_{kl}\exp\Biggl[-\gamma\Bigl(\sum_{(a,b)\in P_\cdot}x_ax_b+ \sum_{c\in P_\cdot} x_c\Bigr)\Biggr],
\end{equation}
where $\gamma > 0$ is the effect-size parameter. In this model, both correlations and anti-correlations between different brain regions are reduced in magnitude.



The final piece in the imaging genetics simulation engine is the method for assigning a case-control disease status to each subject. This serves the following two purposes. It allows us to establish the power of imaging genetics via a comparison of quantitative vs. case-control RF regressions, and it is used to validate the supervised learning approach to define distances between brain networks. 

To minimise the model complexity, we only consider the effects of the increase in vectorial phenotype on the disease classification, neglecting the contribution of the decrease in brain connectivity. Given that a correlation between structural and functional connectivity is built in into our simulations, this is not an unreasonable simplification. 

Based on empirical observations of the ADNI data, we have the following design considerations for constructing an AD disease classification model. Firstly, such a model is inherently probabilistic with the propensity to develop AD increasing with the overall ventricular volume gradient of a sub-region of the whole brain. Analysis of the ADNI data in \citep{silver} suggests that this highly discriminating region is made up of $\sim7\%$ of the total brain area; this corresponds, in our simulation set-up of 400 parcels, to 28 ROI. Secondly, as already mentioned, the distribution of volume gradients in the output of such a model, for both AD and CN classes, must agree with the empirical conditional probability distributions from the ADNI data. Thirdly, the model should allow for a range AD population penetrance values which can be set independently of the effect size. We propose, therefore, the following simple, semi-empirical, Bayesian probability model for AD classification based on the average gradient of the causal region  $R_\text{d}$.

Let the $\bar{{y}}_{(\text{d})i}^*$ represent the average gradient of the causal region $R_\text{d}$ for a given subject $i$, i.e.
\begin{equation}
\bar{y}^*_{(\text{d})i} = \frac{1}{n_\text{d}}\sum_{k\in R_\text{d}} y_{ik}^*,
\end{equation}
where $n_\text{d}=28$ in this simulation set-up, and the $*$ symbolises the values post-genetic effects. Let the disease status be wholly determined by $\bar{{y}}_{(\text{d})i}^*$, i.e. we have the following equivalence between conditional probabilities for a subject $i$ begin given an AD classification
\begin{equation}
\prob(z_i =1 \mid \mathbf{y}_i^*, \Sigma_{kl}^*) \equiv \prob(z_i=1 \mid \bar{{y}}_{(\text{d})i}^*),
\end{equation}
where $z_i =1,0$ are the AD/CN classification label for subject $i$ respectively. Using Bayes's Theorem, we write
\begin{equation}\label{bayespos}
\begin{split}
\prob(z_i=1 \mid \bar{{y}}_{(\text{d})i}^*) &= \frac{\prob(\bar{{y}}_{(\text{d})i}^*\mid z_i=1) \prob(z_i=1)}{\prob(\bar{{y}}_{(\text{d})i}^*)}\\
 &= \frac{\prob(\bar{{y}}_{(\text{d})i}^*\mid z_i=1) \prob(z_i=1)}{\prob(\bar{{y}}_{(\text{d})i}^*\mid z_i=1)\prob(z_i=1) + \prob(\bar{{y}}_{(\text{d})i}^*\mid z_i=0)(1 - \prob(z_i=1))},
\end{split}
\end{equation}
where $\prob(z_i=1)$ is the AD population penetrance level. The likelihoods $\prob(\bar{{y}}_{(\text{d})i}^*\mid z_i=0)$ and $\prob(\bar{{y}}_{(\text{d})i}^*\mid z_i=1)$ are empirically determined from the ADNI data. As before, we assume a Gaussian distribution in the ADNI data, with the sample means $\hat{\mu}_\text{CN} =-0.38$ and $\hat{\mu}_\text{AD} =0.57$ for the CN and AD classes respectively. To avoid the situation where the probability of a AD classification given a large, negative, $\bar{{y}}_{(\text{d})i}^*$ is greater than the probability of a CN classification, and vice versa, we assume that the variances of the normal distributions, representing the respective likelihoods of the CN/AD classes, are identical. We therefore take the average of the two standard deviations, i.e. $\hat{\sigma} = \frac{1}{2} (\hat{\sigma}_\text{CN} +\hat{\sigma}_\text{AD})=0.90$. In Figure \ref{bayespos_fig} we show the empirically determined AD classification probability \eqref{bayespos} as a function of $\bar{{y}}_{(\text{d})i}^*$ for two different AD population penetrance levels (0.20 and 0.35).

\begin{figure}[ht] 
\centering
{\includegraphics{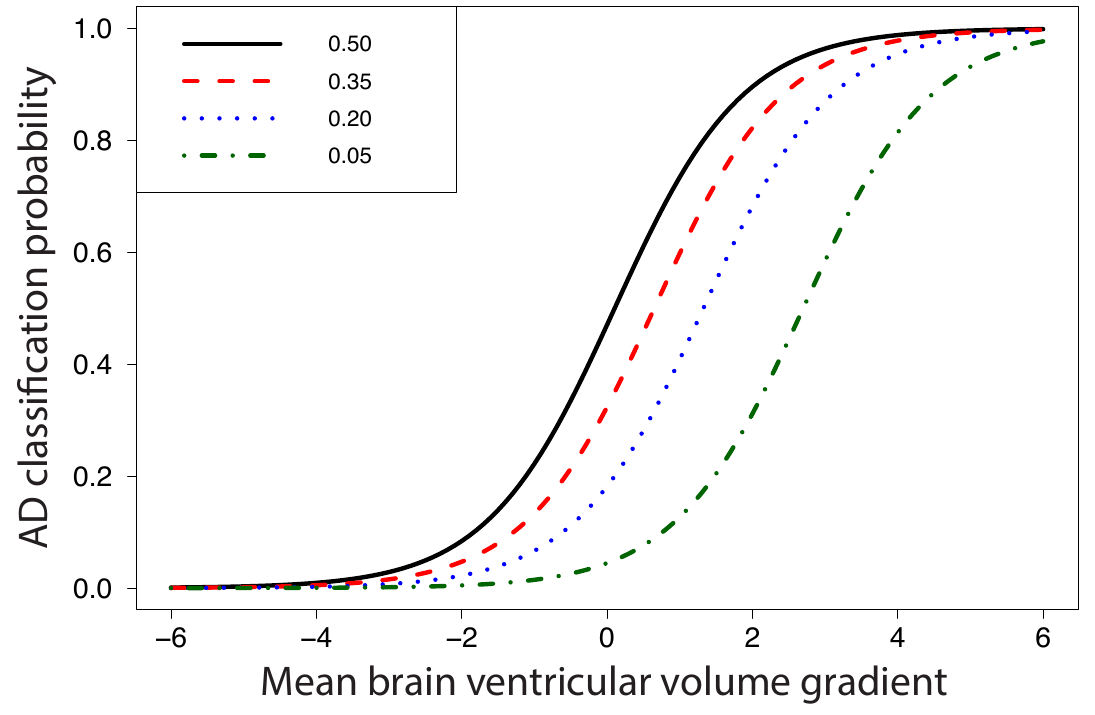}} 
\caption[AD disease classification probability]{The empirical AD classification probability as a function of mean brain ventricular volume gradient. The individual curves (green, blue, red, and black) represent a range of AD population penetrance levels (5\%, 20\%, 35\%, and 50\% respectively).}
\label{bayespos_fig}
\end{figure}

\bibliographystyle{DeGruyter}
\bibliography{distRF5}

\end{document}